 \documentclass[pmlr,twocolumn,10pt]{jmlr} 
\usepackage{wrapfig} 

\usepackage{placeins}

\usepackage{booktabs}
\usepackage{siunitx}

\usepackage[switch]{lineno}

\newcommand{\equal}[1]{{\hypersetup{linkcolor=black}\thanks{#1}}}

\theorembodyfont{\upshape}
\theoremheaderfont{\scshape}
\theorempostheader{:}
\theoremsep{\newline}

\jmlrvolume{259}
\jmlryear{2024}
\jmlrsubmitted{LEAVE UNSET}
\jmlrpublished{LEAVE UNSET}
\jmlrworkshop{Machine Learning for Health (ML4H) 2024} 

\title[
Learning
Personalized
Treatment 
Decisions 
in 
Precision Medicine
]{
Learning 
Personalized 
Treatment 
Decisions 
in Precision Medicine:
Disentangling 
Treatment 
Assignment
Bias
in 
Counterfactual 
Outcome
Prediction
and
Biomarker Identification
}

\author{%
\Name{Michael Vollenweider}\equal{Equal contribution} \Email{michavol@ethz.ch}\\
\addr ETH Zurich
\AND
\Name{Manuel Schürch}\footnotemark[1] \Email{manuel\_schurch@dfci.harvard.edu}\\
\addr University of Zurich, University Hospital Zurich, Harvard University
\AND
\Name{Chiara Rohrer}
\Email{rohrechi@student.ethz.ch}\\
\addr ETH Zurich
\AND
\Name{Gabriele Gut}
\Email{gabriele.gut@usz.ch}\\
\addr University of Zurich, University Hospital Zurich 
\AND
\Name{Michael Krauthammer}
\Email{michael.krauthammer@uzh.ch}\\
\addr University of Zurich, University Hospital Zurich 
\AND
\Name{Andreas Wicki}
\Email{andreas.wicki@usz.ch}\\
\addr University of Zurich, University Hospital Zurich 
}

\begin{document}

\maketitle

\begin{abstract}
Precision medicine has the potential to tailor treatment decisions to individual patients using machine learning (ML) and artificial intelligence (AI), but it faces significant challenges due to complex biases in clinical observational data and the high-dimensional nature of biological data.
This study models various types of 
treatment assignment
biases using mutual information and investigates their impact on 
ML
models for counterfactual prediction and biomarker identification. Unlike traditional counterfactual benchmarks that rely on fixed treatment policies, our work 
focuses on 
modeling 
different characteristics of the underlying observational treatment
policy 
in
distinct clinical settings. We validate our approach through experiments on toy datasets, semi-synthetic tumor cancer genome atlas (TCGA) data, and real-world biological outcomes from drug and CRISPR screens. By incorporating empirical biological mechanisms, we create a more realistic benchmark that reflects the complexities of real-world data. Our analysis reveals that different biases
lead to varying model performances, with some biases, especially those unrelated to outcome mechanisms, having minimal effect on prediction accuracy. This highlights the crucial need to account for specific biases 
in 
clinical 
observational data 
in 
counterfactual ML model development, ultimately enhancing the personalization of treatment decisions in precision medicine.
\end{abstract}

\paragraph*{Data and Code Availability.}
The datasets 
are either synthetically generated or 
can be downloaded from publicly available sources.
The source code is available on \href{https://github.com/michavol/selection-bias-benchmark}{GitHub}.


\paragraph*{Institutional Review Board (IRB).}
No IRB approval was required for this study.
%

\section{Introduction}
The application of machine learning (ML) 
and artificial intelligence (AI) to observational data offers significant potential for advancing precision medicine. In particular, domains such as personalized diagnostics and prognosis as well as therapy decision support are poised to be primary beneficiaries. To make progress on the latter, much effort has been put into transitioning from learning average treatment effects (ATE) to personalized treatment effects, 
referred to as conditional average treatment effects (CATE) \citep{bica2021real, curth2024using}. A critical challenge in this domain remains 
\textit{counterfactual} prediction — the task of determining what would have happened had a different action been taken instead of the 
\textit{factual}
one. 


\begin{figure}[t!]
    \centering
\includegraphics[width=0.49\textwidth]{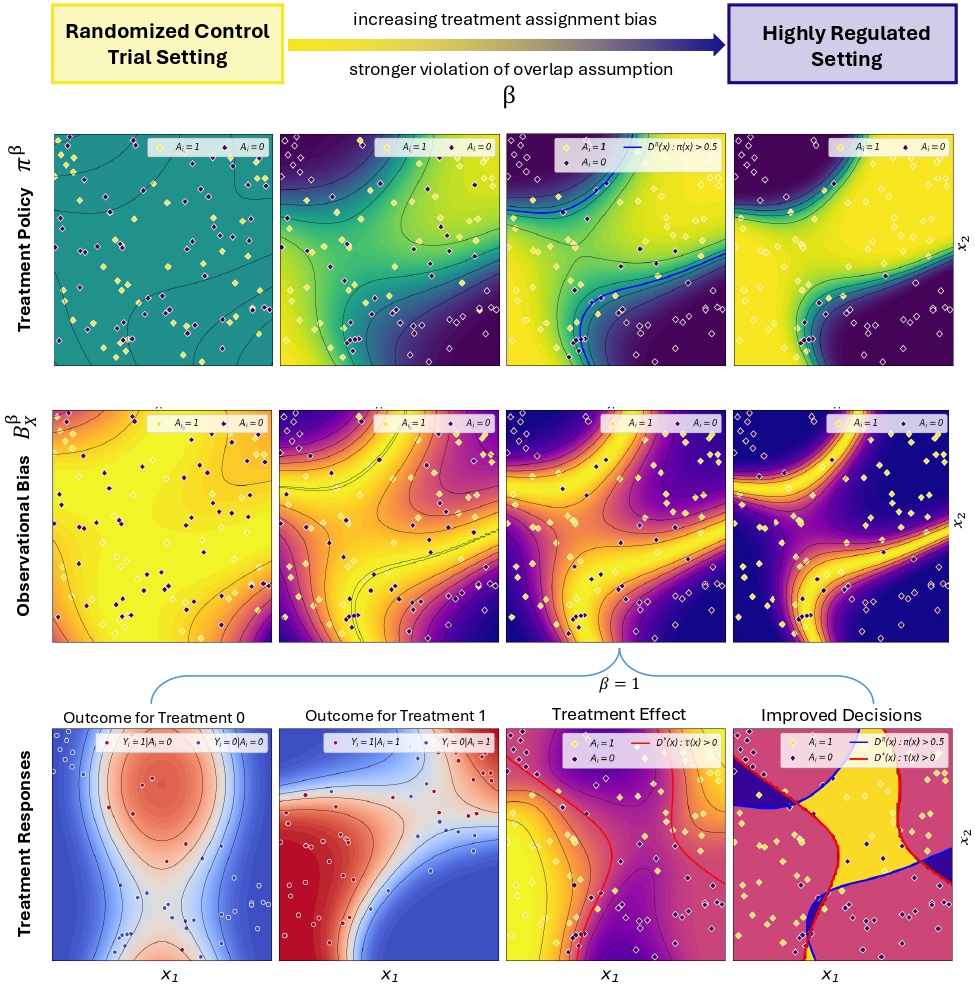} 
    \caption{
    \small{
   Different treatment assignment policies 
   with increasing bias
   for simplified patient data represented by ($x_1$, $x_2$). 
   Row 1: 
   Lighter regions indicate higher likelihood for patients of receiving treatment 
   1.
   Row 2: Darker areas show strong treatment selection bias. 
   Row 3: Estimating the outcome surfaces for both treatments and the conditional treatment effect helps to refine policies to optimize outcomes, as seen with the current suboptimal treatments in the blue and yellow regions.
    %
    }
    }
\label{fig:fig1}
\end{figure}

\noindent Randomized Controlled Trials (RCTs) 
are
considered the gold standard for 
supporting
treatment decision
in healthcare
because they lack treatment selection bias, which arises when treatment assignment is biased toward specific patient characteristics and generally complicates prediction \citep{hariton2018randomised}. However, RCTs face limitations, such as high costs, ethical concerns, 
and challenges in generalizing findings to broader populations.
Moreover, in the context of high-dimensional data,
it is impossible to control for all variables.
On the other hand,
the abundant and diverse nature of retrospective observational data offers promising opportunities for learning personalized 
treatment decisions
with ML models.
However, the treatment selection bias present in such data can lead to skewed or unrepresentative predictions.

%

%
%
\paragraph{Clinical Relevance of Bias.}
Different clinical settings exhibit specific types of bias. For example, treatment selection can be biased towards cost in low-income settings and treatment effectiveness in high-income settings. Bias may also occur because therapy decision-making is informed by a flawed study and, consequently, treatments are assigned based on incorrectly identified biomarkers. Furthermore, even if the study is not flawed, treatment decisions are almost always biased toward the average effect. 
The bias of an observed treatment policy, that is, the protocol used to administer care to patients, may also change throughout the course of patient treatment. This is especially noticeable in cancer care, where initial treatment policies are often strictly regulated by medical guidelines (informed through clinical trial evidence), leading to substantial treatment selection bias. As a patient's disease progresses, treatment options are less regulated and policies are more dependent on the expertise of treating physicians and diagnostic evidence. \figureref{fig:fig1} visualizes this idea of varying selection bias and shows how increasing bias results in a more severe covariate shift between treatment groups, complicating counterfactual prediction. 
Therefore, it is crucial to distinguish different clinical settings, and
formalize 
varying notions of bias as it will help analyze their effects for counterfactual prediction and biomarker identification.

\begin{figure}[t!]
    \centering
\includegraphics[width=0.5\textwidth]{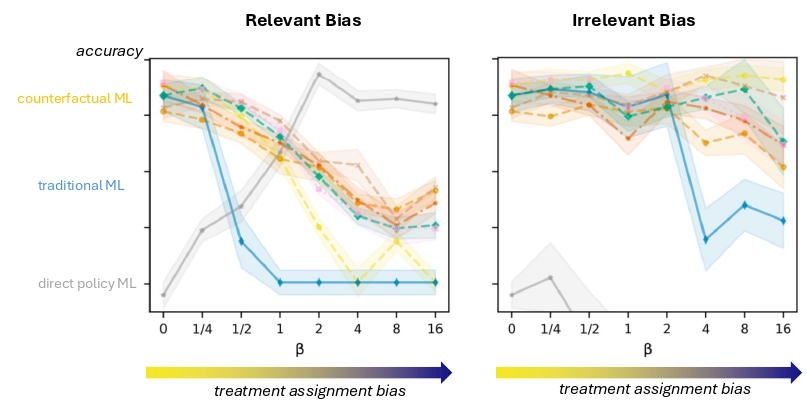} 
    \caption{
    \small{
Key findings in this paper:
1) Addressing various treatment assignment biases under the consideration of their strength $\beta$ and relevance, is essential
when learning from high-dimensional observational data.
2) Counterfactual ML approaches can enhance estimating optimal treatment policies compared to traditional ML; however, there is still room for improvement in effectively utilizing relevant biases in observational treatment policies.
    }
    }
\label{fig:fig1b}
\end{figure}

\noindent\textbf{Violation of Overlap Assumption.}
Substantial research has focused on defining the theoretical assumptions necessary for counterfactual prediction and developing methods to handle selection bias in observational data. A key assumption in this context is the overlap assumption \citep{rosenbaum1983central}, which ensures that each patient has a non-zero probability of receiving any treatment. Observational policies with strong selection bias can violate this assumption. In precision medicine, data are often high-dimensional, that is, there are more features than samples. In this case, the overlap assumption is basically never fulfilled \citep{d2021overlap}. Is there hope to learn from this high-dimensional observational data, nevertheless? We argue that the violation of the overlap assumption does not necessarily harm prediction performance and that it is crucial to differentiate between various types of bias, as they can have distinct effects. To our knowledge, different kinds of observational policies have only enjoyed little investigation. Benchmarks often simulate different outcomes but usually fix selection policies without varying the type of bias \citep{louizos2017causal, johansson2016learning, shi2019adapting}.
However, there are fundamental differences between learning from observational data in a setting close to a randomized controlled trial (RCT) and in a highly regulated environment, 
as illustrated in \figureref{fig:fig1},
since this significantly impacts the choice and performance of counterfactual ML models. \figureref{fig:fig1b} summarizes 
our key 
findings for different characteristics of the underlying treatment policy and different ML models.

\paragraph{Evaluation on Biological Outcomes.}
The effects of bias also depend on the data-generating processes that underlie the observational outcomes. Often, treatment effect models are evaluated on semi-synthetic datasets, which 
often favor certain classes of models because of the specific underlying data-generating process. Following the recommendations of \citet{gentzel2019caseevaluatingcausalmodels, schurch2024towards}, in addition to semi-synthetic data, we propose using empirical potential outcomes derived from in-vitro cell line experiments from DepMap \citep{Tsherniak2017} and simulating treatment assignment policies with certain characteristics. This allows for a more robust evaluation by incorporating real-world biological variability and complexities that are often absent in purely synthetic datasets. 


\paragraph{Related Work.} 
The focus of this work is on different characteristics of  
treatment policies, which makes it different from the several benchmarks for 
counterfactual 
predictions,
which
usually simulate different outcomes, but fix an arbitrary selection policies.
Furthermore, in addition to CATE prediction, we also study the effect of different policy characteristics on the prediction of counterfactual outcomes and biomarker identification \citep{crabbé2022benchmarkingheterogeneoustreatmenteffect}. 
Our formalization of bias adapts and extends the concept of expertise, introduced by \citet{definingExpertise}. While the introduction of expertise provides valuable insights, the given interpretation does not fully capture the nuances of selection bias that we are interested in, and the results on how bias affects absolute prediction performance differ significantly from ours. 
Furthermore, as pointed out by \citet{gentzel2019caseevaluatingcausalmodels} and \citet{curth2021really}, related work in this area suffers from the lack of biologically realistic evaluation datasets and often relies only on simulated results.

\paragraph{Contributions.} We formalize and quantify various types of treatment assignment biases
induced 
by observational treatment policies and explain how they relate to different clinical settings and types of biomarkers. 
We systematically simulate different types of 
treatment selection policies and analyze their effects on the performance 
of various state-of-the-art counterfactual ML models
when combined with toy, semi-synthetic, and real outcomes. 
We propose employing in-vitro 
experiments for counterfactual evaluation, thereby offering the community a novel evaluation approach characterized by realistic outcomes and multi-modal biological covariates. The results demonstrate that the type of bias matters significantly, that the violation of the overlap assumption is not necessarily detrimental, and that models differ in how they respond to various biases leading to several important findings for developing new methodologies and algorithms of counterfactual treatment models tailored to different settings in
precision medicine.

\label{sec:related_works}


\section{Methodology}

\subsection{Data Generating Process (DGP)}

\begin{figure}[t]
    \centering
\includegraphics[width=0.42\textwidth]{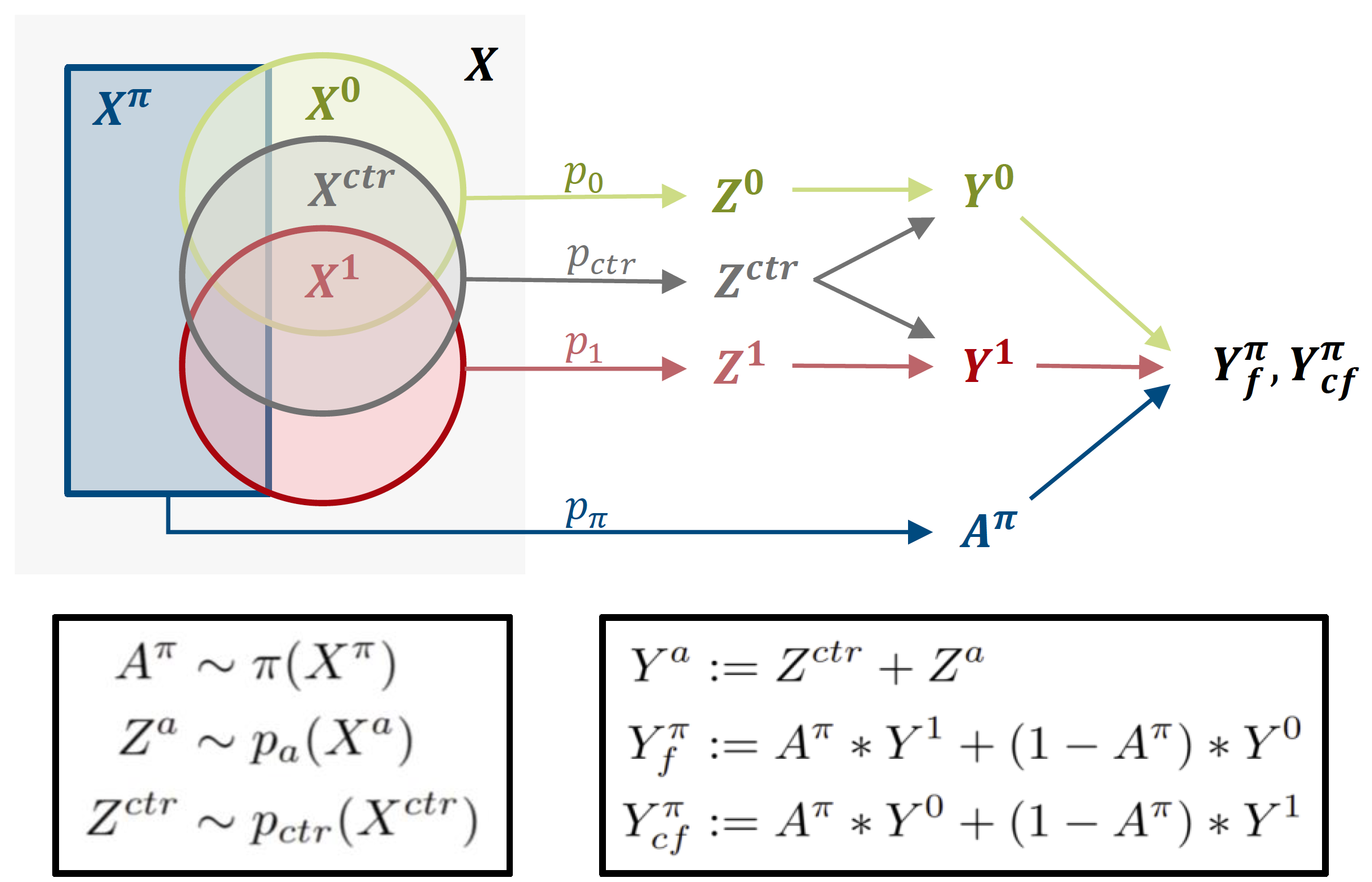} 
    \caption{Data Generating Process (DGP).}
    \label{fig:fig2}
\end{figure}

We consider 
patient characteristics
$X\in \mathcal{X}$, binary treatment assignments $A^{\pi}\in 
\{0,1\}$ under a particular treatment assignment policy $\pi$, and 
real-valued
treatment outcomes $Y\in 
\mathbb{R}
$.
These variables are  linked together 
and 
the underlying probabilistic data-generating process 
can be defined 
through a structural causal model (SCM) \citep{Peters2017}. 
As 
defined
in \figureref{fig:fig2},
%
%
we consider a 
treatment assignment policy
$\pi(X^\pi)$, representing
the probability 
that a patient 
gets the first treatment $A^\pi=1$ 
under
the observational policy $\pi$.
%
Moreover,
 we partition $X$ into various subsets; 
the selective features
$X^\pi \subseteq X$ which
 include all features influencing the treatment assignment, 
the treatment-specific  sets 
$X^{0}, X^{1}\subseteq X$,
and the treatment-independent control 
 set $X^{ctr}\subseteq X$.
This allows to define detailed treatment outcome mechanisms for the potential outcomes
$Y^0$ and $Y^1$ depending on different subset of $X$ and is useful to benchmark different biomarker identification.
Depending on the treatment assignment, the factual outcome $Y^\pi_f$ corresponds to one of the potential outcomes $Y^0$ or $Y^1$. However,
the fundamental problem of
counterfactual inference \citep{rubin2005} from observational data is that never both potential outcomes are observed together for a particular patient, that is,
only the
factual $Y^\pi_f$ but never the counterfactual $Y^\pi_{cf}$,
making this task 
challenging.

\subsection{Quantities of Interest} \label{subsec:quantities}
\paragraph{Potential Outcomes (POs).
} 
The DGP 
characterizes the 
potential outcomes 
$Y_i^0$ and $Y_i^1$ for each patient $i$,
as well as the 
conditional
expected potential outcome \citep{rubin2005}
\[\mu_a(x) = E[Y_i^a\,|\, X_i=x].\]
For an oncological observational dataset, the potential outcomes 
$Y^0$ and $Y^1$ could represent clinically measured tumor sizes or the progression-free survival under standard care versus a new chemotherapy drug, respectively. 
Knowing this quantity helps clinicians understand how patients react to certain treatments, based on their specific characteristics.

\paragraph{Treatment Effects.}
The difference $\tau_i=Y_i^1-Y_i^0$ is the individual treatment effect and quantifies to what extent one treatment option is preferable over another.
The expectation of 
$\tau_i$
leads to the
 conditional average treatment effect (CATE) \citep{rubin2005}
\[\tau(x) = E[Y_i^1 - Y_i^0 \,|\, X_i=x] = \mu_1(x) - \mu_0(x).\]
%
Since the CATE is a deterministic function in both expected potential outcomes, it captures strictly less information but is sometimes easier to estimate.

\paragraph{Optimal Treatment Policies.} Ultimately, a physician aims to make the best possible decisions for every new patient. If the objective is to find a policy that always assigns the treatment with the greater outcome, i.e. that $Y^\pi_f(x) \geq Y^\pi_{cf}(x)$ for all patients, then an optimal deterministic decision policy can be 
defined as
$D^*(x) = \arg\max_{a \in \{0,1\}} Y^a(x)$, which is
illustrated in Fig.\ \ref{fig:fig1} in the last row/column.
Note that in some clinical settings, the objective may not be as simple as maximizing the outcomes \citep{li2023optimal, schurch2023generating}. One may also consider more risk-aware objectives or joint optimization of multiple outcomes.  

\begin{figure}[t]
    \centering
\includegraphics[width=0.5\textwidth]{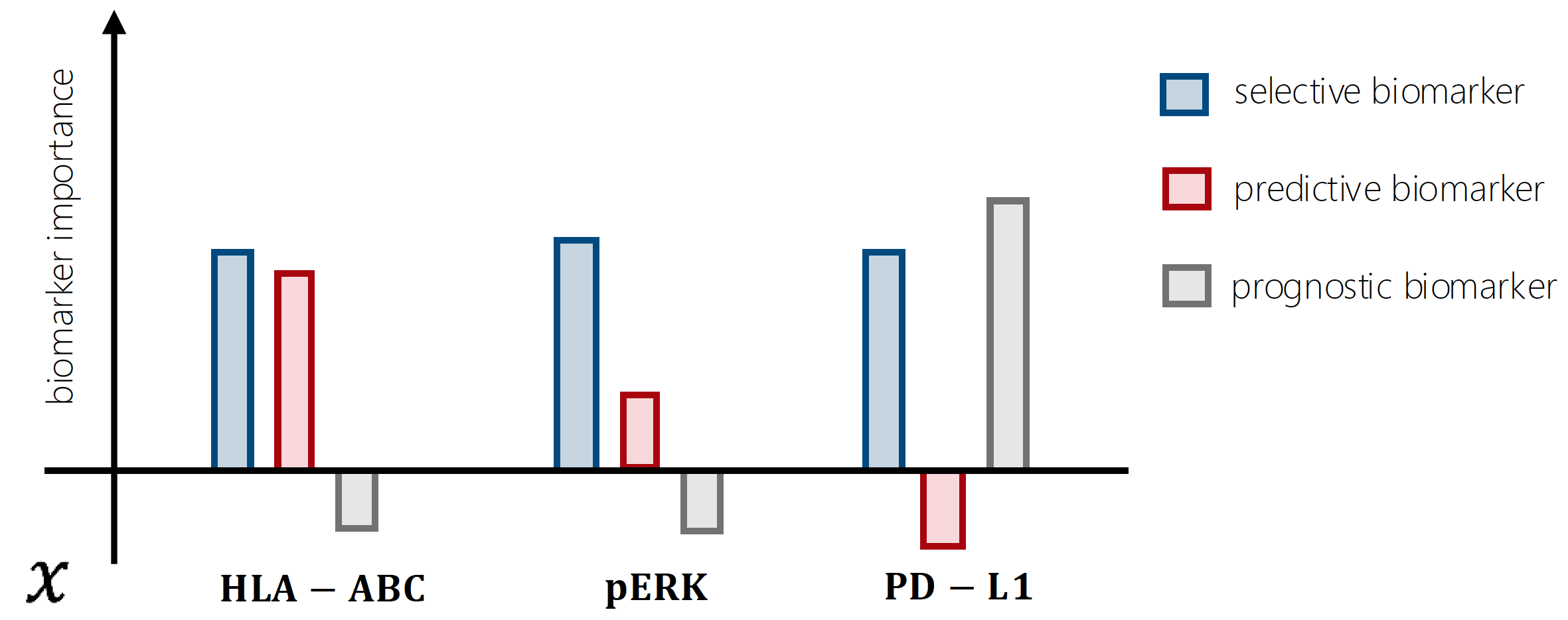} 
    \caption{Different clinical biomarker.
    }
    \label{fig:fig3}
\end{figure}

\paragraph{Clinical Biomarkers.} 
In healthcare, biomarkers play a crucial role in guiding treatment decisions. We distinguish between \textit{prognostic} biomarkers, which predict the overall disease outcome regardless of the treatment, and \textit{predictive} biomarkers, which indicate how likely a specific treatment will be effective, following 
\citet{sechidis2018distinguishing, crabbé2022benchmarkingheterogeneoustreatmenteffect, garnett2012systematic}.
Additionally, we introduce the concept of \textit{selective} biomarkers — those biomarkers that clinicians actively use to assign treatments.
 As illustrated in the fictional example in \figureref{fig:fig3}, we can assess the importance of different biomarkers through feature attribution methods \citep{lundberg2017unified}. 
 Ideally, selective biomarkers, for instance, HLA-ABC, align with predictive biomarkers.
 On the other hand,
 the PD-L1 biomarker is identified as both prognostic and selective, meaning it influences decision-making but primarily affects disease outcome rather than treatment effectiveness. This suggests PD-L1 might not be the best guide for treatment selection. 
 Further, the example of pERK illustrates a biomarker that is used to make treatment decision but is neither a predictive nor prognostic feature.
 By understanding the roles of these biomarkers, we can better evaluate and improve the decision-making process in clinical settings. The DGP in \figureref{fig:fig2} allows 
 to simulate different ground-truth feature sets and importances to validate the accuracy of biomarker discovery.

\subsection{
Quantifying
Treatment
Assignment Bias
} \label{subsec:biases}
To be able to analyze 
different 
treatment assignment policies
$A^{\pi}\sim\pi(x) = P(A^{\pi} = 1 \mid X = x)$, we quantify 
different treatment assignment biases 
based on 
mutual 
information (e.g. \cite{murphy2022probabilistic}).
%
\begin{definition}[Z-Bias] The Z-bias for any treatment assignment $A^{\pi}\sim \pi(X)$ and random variable $Z$ is defined as
\begin{equation} \label{eq:bias}
    B^\pi_{Z} = \frac{\mathbb{I}(A^\pi;Z)}{\mathbb{H}[A^{\pi}]},
\end{equation}
which measures the degree 
of bias
of
treatment assignment $A^{\pi}$ 
with respect to 
$Z$.
\end{definition}
The Z-bias is proportional to the 
mutual information $\mathbb{I}(A^{\pi}, Z)=\sum_{a \in \{0,1\}} \int_{-\infty}^{\infty} p(a, z) \log \left( \frac{ p(a, z) }{ p(a) p(z) } \right) dz$ 
between treatment assignment $A^{\pi}$ and random variable $Z$ and quantifies the amount of information provided by $Z$ about the treatment assignment. It is normalized by the marginal entropy of treatment assignments $A^{\pi}$, that is $\mathbb{H}[A^{\pi}] = -\sum_{a \in \{0,1\}} p(a) \log(p(a))$ 
and 
measures the general uncertainty in treatment assignment.  $B^\pi_{Z}$ attains its minimum and maximum at 0 and 1, respectively. In this way, $Z$-bias may also be interpreted as the portion of variability in treatment assignment that can be explained by $Z$. If there is no association between the assignment of the treatment and $Z$, then $B^\pi_{Z}=0$ (no bias). However, if the policy is a deterministic function of $Z$, knowing $Z$ fully explains all treatment assignments and $B^\pi_{Z}=1$ (high bias). There are five important special cases which we will use for our experiments. 
\\
$
\mathbf{
B_{X}^\pi}$.
Observable bias or $X$-bias defined as $B_{X}^\pi
=
\mathbb{I}(A^\pi;X)/
\mathbb{H}[A^{\pi}]$ quantifies how much information all patient characteristics provide about treatment selection, as depicted in the second row in Fig.\ \ref{fig:fig1}. This bias can also be seen to quantify the degree of violation of the overlap assumption in causal inference. The assumption ensures that every patient has a non-zero probability of receiving each treatment level given their covariates, that is,
$0 < \pi(x) < 1  \text{ for all } x \in \mathcal{X}$ \citep{rubin2005}. 
A policy $\pi$ which deterministically assigns treatments based on $X$, fully violates this assumption and attains the maximal $X$-bias of 1. 
\\
$\mathbf{B^\pi_{Y^0}} \textbf{ \& } \mathbf{B^\pi_{Y^1}}$. The treatment outcome biases, $Y^0$-bias and $Y^1$-bias, 
defined as
$B_{Y^a}^\pi
=
\mathbb{I}(A^\pi;Y^a)/
\mathbb{H}[A^{\pi}]$, respectively, 
describe to what extent the potential outcomes of the treatments $a=0$ and $a=1$ determine the treatment decision. If treatment $a=0$ corresponds to a control (no treatment), then $B^\pi_{Y^0}$ could be termed prognostic bias, as the control outcome is determined by prognostic biomarkers.
\\
$\mathbf{B^\pi_{Y^1-Y^0}}$. The treatment effect bias defined as
$B_{Y^1-Y^0}^\pi
=
\mathbb{I}(A^\pi;Y^1-Y^0)/
\mathbb{H}[A^{\pi}]$
quantifies how informed the assignment policy is with respect to the true treatment effect. Similarly to $Y^0$-bias, if treatment $a$ constitutes a control, then $B^\pi_{Y^1-Y^0}$ may also be called predictive bias, as the treatment effect is determined by predictive features. 
\\
$\mathbf{B^\pi_{Y^0, Y^1}}$.
The total outcome bias defined as
$B_{Y^0, Y^1}^\pi
=
\mathbb{I}(A^\pi;Y^0, Y^1)/
\mathbb{H}[A^{\pi}]$
describes to what extent the joint distribution of the potential outcomes determines the treatment decision. 
\\
\noindent
The special cases of $Z$-bias, $B_{Y^1-Y^0}^\pi$ and $B^\pi_{Y^0, Y^1}$ 
are closely related to 
 predictive and prognostic expertise, respectively,
 introduced
 by \citet{definingExpertise}. 
 In addition, $B_{X}^\pi$ is similar to their concept of in-context variability. While some of the definitions 
 in this paper 
 are
 related to
 their work, our interpretation and some results differ significantly. See \appendixref{apd:bias_discussion} for a discussion of why we prefer $Z$-bias over $Z$-expertise 
 to describe the quantity in \equationref{eq:bias}. 


\begin{figure}[t]
\centering    \includegraphics[width=0.48\textwidth]
    {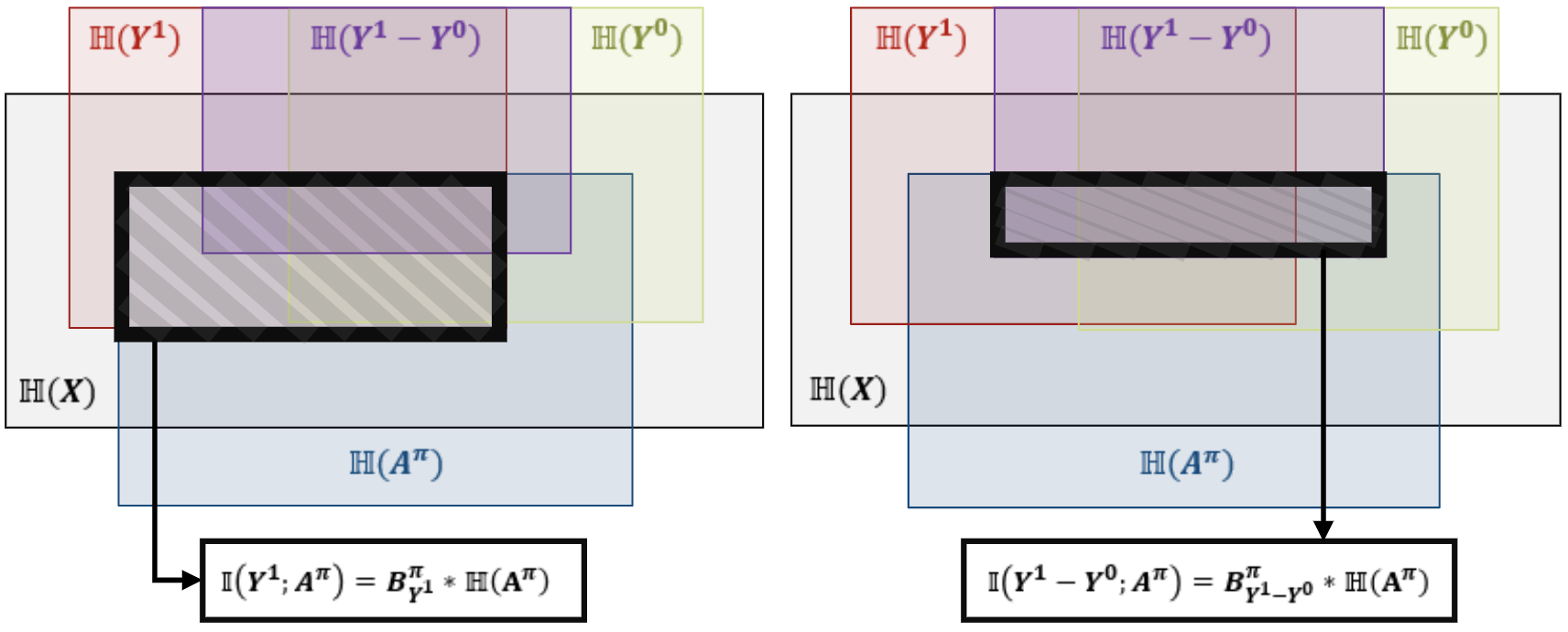}
    \caption{Illustration of different biases.
    }
    \label{fig:fig4}
\end{figure}

\begin{proposition} \label{prop:ineq}
  Under the non-confounding assumption, we have 
\begin{equation} \label{eq:bias_ineq}
\begin{aligned}
    B^\pi_X \geq B^\pi_{Y^0, Y^1} & \geq B^\pi_{Y^0}, B^\pi_{Y^1}, B^\pi_{Y^1-Y^0}. \\
\end{aligned}
\end{equation}
\end{proposition}
%
%
\noindent
Proposition \ref{prop:ineq} establishes an ordering between different types of bias, which is visualized in the Venn diagram for $B_{Y^1}^\pi$ and $B_{Y^1-Y^1}^\pi$ in \figureref{fig:fig4} and 
a proof can be found in \appendixref{apd:proofs}. 

\noindent
We also refer to \figureref{fig:biases} in \appendixref{apd:proofs} for a visualizations of all considered and simulated biases. 
The entropies of random variables are represented as rectangles, and their mutual information as the overlap thereof. The overlaps with the entropy of $A^\pi$, correspond to a representation of the $Z$-biases, scaled by the entropy of $A^\pi$.
The manner in which the entropies relate is generally complex and depends on the distributions arising from the underlying noisy causal structural mechanisms. 

\begin{proposition} \label{prop:oa}
$Z$-bias relates to the overlap assumption (OA) as follows
\begin{equation} \label{eq:oa}
\begin{aligned}
    B^\pi_{X} = 1 \implies \textit{OA is violated.} \\
\end{aligned}
\end{equation}
\end{proposition}
Proposition \ref{prop:oa}
states that when there is maximal obervational bias, then the overlap assumption is necessarily violated. This relates Z-bias to the commonly used assumption used in literature and will allow to measure how strongly it is violated in different simulation settings. Proofs and further results can be found in \appendixref{apd:proofs}.


\paragraph{Simulating Treatment Assignment Policies}
To be able to simulate different kinds of treatment
assignment policies $\pi(x)$ with a controlled amount of bias, we subsequently define a parametrized policy.
\begin{definition}[Z-Policy]
For a given random variable \( Z \)
and parameter \( \beta \), the \( Z \)-policy is defined as 
%
\begin{equation} \label{eq:policy}
\pi_Z^{\beta}(x) 
:= \sigma(\beta Z(x)),
\end{equation}
where 
%
\(
\sigma(x) := \frac{e^x}{1 + e^x}.
\)
This Z-policy represents the probability that a patient with characteristics \( X = x \) is assigned to treatment \( A^{\pi_Z^\beta} = 1 \) under the \( Z \)-policy.
\end{definition}
When $\beta = 0$, the policy mimics a (balanced) RCT with $\pi_Z^{\beta=0}(x) = 1/2$ and hence, $\pi_Z^{\beta=0} = \pi_{RCT}$. As $\beta$ increases, treatment selection becomes more dependent on $Z$, leading to greater selection bias. This parameter $\beta$ is thus termed the bias scale. \figureref{fig:fig1} in the first row visualizes this process: increasing $\beta$ simulates a transition from an unbiased RCT setting to a highly biased observational policy.
%

\section{Experiments}
\subsection{Data}
In total, we use four types of datasets for our policy simulation experiments based on toy examples, TCGA data, CRISPR screens and drug screens. 
With the carefully constructed toy examples, we aim to elucidate key concepts. Their setup and corresponding results are discussed in \appendixref{apd:toy_examples}.
\paragraph{Simulated Outcomes.} To include a more conventionally used dataset, we extend the linear semi-synthetic simulation setting used in \citet{crabbé2022benchmarkingheterogeneoustreatmenteffect} with $n=1000$. Here, the covariates are 100 real RNA transcriptomics data from the TCGA databank.
This dataset will be referred to as AY-TCGA, as we simulate the treatment and outcome mechanism.
See \appendixref{apd:data} for details on the setup.
\paragraph{Biological Outcomes.} To create a more realistic setting, we leveraged in-vitro data from a drug and a CRISPR screen from DepMap (Dependency Map) \citep{Tsherniak2017}. The use of 
in-vitro data
ensures high genetic consistency, with biological outcome measurements for different perturbational experiments,
closely approximating the true counterfactuals. 
We refer to the two datasets derived from DepMap as A-CRISPR and A-DRUG, since we only simulate the treatment selection (blue path associated within our DGP in \figureref{fig:fig2}) but use the 
real biological treatment outcome mechanisms. 
For a more in-depth description 
see \appendixref{apd:data}. 

\subsection{Learners}
Given observational data $\mathcal{D}=\{x_i, a_i, y_i\}_{i=0}^n$, generated with some observed policy $\pi$ according to the DGP in \figureref{fig:fig2}, the goal is to estimate the quantities discussed in Sec.\ \ref{subsec:quantities}. The fundamental difficulty of this task is the absence of measurements of the counterfactual outcome $Y_{cf}^\pi$.
There exist several state-of-the-art counterfactual methods that aim to deal with this difficulty while estimating the quantities $\hat{\mu}^a(x), \hat{\tau}(x)$ and $\hat{\pi}(x)$. For comparison in our experiments, we selected models based on three primary axes: linear vs. nonlinear, action-predictive vs. balancing, and direct vs. indirect approaches. These distinctions guide our understanding of how various model designs handle treatment effect prediction and interact with different types of biases inherent in observational data.
\noindent
We use the 
implementations of SLearner, TLearner, XLearner, CFRNet, DragonNet, and TARNet provided by \citet{curth2021nonparametric}, and the EconML implementations of SLearner and TLearner \citep{econml}. See \appendixref{apd:learners} for further details on three primary axes and a description of the selected learners.
We also introduce ActionNet, which trains a propensity net on observational treatment decisions and assigns $A^{\hat{\pi}}=0$ if the predicted propensity is less than 0.5 and $A^{\hat{\pi}}=1$, otherwise. 
To assess the consistency of model performance, the models were trained on five random seeds and five-fold splits of the data.

\subsection{Evaluation Metrics}  
For treatment effect prediction we employ the classically used \textit{Precision in Estimation of Heterogeneous Effect (PEHE)} and the assignemnt precision metric Prec$_{Ass.}^\pi$, counting how frequently the observational policy $\pi$ selects the optimal outcome, inspired by \citet{curth2021really}. For evaluating potential outcome (PO) prediction, we use the \textit{Root Mean Squared Error (RMSE)} for both factual and counterfactual prediction. To quantify the quality of biomarker identification, we adopt the attribution metrics Attr$_{pred}$ and Attr$_{prog}$ used in \citet{crabbé2022benchmarkingheterogeneoustreatmenteffect}, which count the number of correct predictive and prognostic features were selected as biomarkers compared to the ground-truth. To compute the biases involving the mutual information and entropies,
we approximate them by binning continuous outcomes,
similarly to 
\citet{definingExpertise}.  See \appendixref{apd:metrics} for details.


\section{Results}\label{sec:results}

\begin{figure*}[ht]
    \centering
    \includegraphics[width=.97\textwidth]{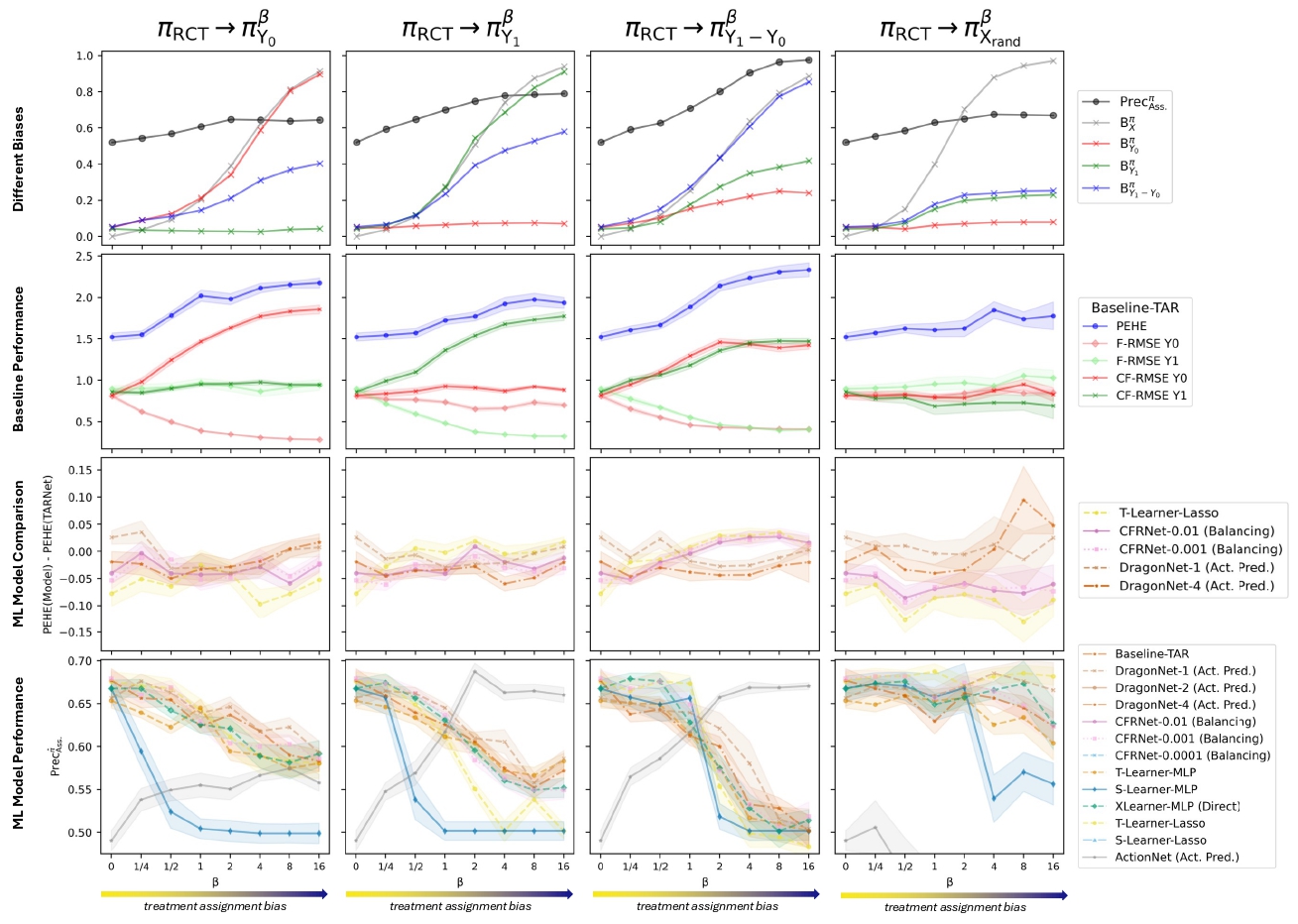} 
    \caption{Results for A-CRISPR.}
    \label{fig:crispr}
\end{figure*}

\figureref{fig:crispr} shows results for the A-CRISPR dataset.
We performed four  experiments 
simulating policies going from an RCT setting ($\beta = 0$) to a fully biased one ($\beta = 16$) with respect to $Y^0, Y^1, Y^1-Y^0$ and $X_{rand}$. Here, $X_{rand}$ represents a weighted linear combination of 20 randomly chosen features and uniformly sampled weights in $[-1,1]$, imitating a situation where the treatment decision is misinformed. By normalizing all the quantities before providing them as input to the sigmoid function $\sigma$ in Eq.\ \ref{eq:policy}, we ensure that treatment assignment is balanced. \figureref{fig:tcga} in \appendixref{apd:tcga} shows the results for the same experiments on AY-TCGA with fully simulated outcomes.
In this setting, we can simulate an additional policy $\pi_{X_{pred}}^\beta$, where $X_{pred}$ is a weighted linear combination of 20 predictive features, which determine the treatment effect in the outcome simulation. The weights are also uniformly sampled,
but different from the weights sampled for computing the outcomes. This represents the case where the correct features are used for treatment decisions, but with a wrong/random influence. The results for A-Drug can be found in \appendixref{apd:tcga}.
\paragraph{Biases.} The first row of the plots shows how increasing $\beta$ affects the biases of the obervational policy. 
The observational bias $B_X^\pi$, indicated by the grey line, almost reaches 1 in every setting. 
With Eq.\ \ref{eq:oa}, this means that the overlap assumption becomes increasingly violated with increasing $\beta$. We can also see that for A-CRISPR, the increase of the outcome biases $B^\beta_{Y^0}$ and 
$B^\beta_{Y^1}$ comes with a significant increase of the treatment 
effect bias $B^\beta_{Y^1-Y^0}$. 
This is not the case for simulated outcomes in AY-TCGA, which may be due to correlations between the outcomes, treatment effect and treatment assignment. We believe that the interaction information $\mathbb{I}(A^\pi; Y^0; Y^1-Y^0)$, which is respresented by the overlap of the entropies of all three variables in 
\figureref{fig:fig4}, relates to how much the $Y^0$-bias is linked to the effect bias and consequently, how much they increase together. This difference provides another reason to be cautious when interpreting results from fully simulated outcomes. For the last column, for A-CRISPR, the $Y^1$-bias and treatment effect bias increase slightly. This may indicate that the randomly chosen features, by chance, capture some information about the outcomes.
%


\paragraph{POs \& Treatment Effects.} The second row in Figures \ref{fig:crispr},\ref{fig:tcga}, and \ref{fig:drug}, respectively, depicts how well the baseline model TARNet can predict potential outcomes and treatment effects. 
The results show that the bias for a certain outcome comes with improved factual prediction but worse counterfactual prediction. If there is no outcome bias, there is little to no effect on prediction performance. Similarly, we can see, that the degradation of performance in treatment effect prediction (PEHE)
correlates with the degree of treatment effect bias. This behavior is consistent across all datasets and supports the observation, that an irrelevant violation of the overlap assumption does not necessarily negatively impact prediction, which is a positive message in the context of high-dimensional medical data. The factual and counterfactual RMSE in A-CRISPR also nicely demonstrate the effects of the covariate shift. With increasing bias, the support of the factual part of the outcome surface becomes smaller and thereby the variance of its value range is reduced, making it easier to learn. The counterfactual part of the outcome surface, however, becomes 
increasingly
underrepresented
in the 
training data
and 
the model fails to extrapolate correctly to those regions. The latter effect tends to be stronger, 
hence bias generally leads to worse performance. For the simulated outcomes in AY-TCGA, the factual RMSE does not improve with increasing $\beta$, probably because the outcome surface is easy to learn in the RCT setting. 
In \appendixref{apd:toy_examples},
we show also an example where outcome and effect bias have different effects for direct vs. indirect learners.

\paragraph{Inductive Bias.} If we look at the last row in Fig.\  \figureref{fig:crispr} 
for $\pi_{\text{RCT}} \rightarrow \pi_{Y^1-Y^0}^\beta$, we can see that with high bias, the observational policy makes all the right decisions. In that case, the treatment assignments in the training data themselves provide inductive bias that some models may be able to exploit.
Note that also an observational policy which always decides incorrectly has inductive bias. Indeed, for A-CRISPR, it appears that the action predictive model DragonNet has an advantage over the balancing CFRNet and baseline TARNet for medium to high treatment effect bias. Also for $\pi_{\text{RCT}} \rightarrow \pi_{Y1}^\beta$, where the effect biases increase significantly with the outcome bias, the DragonNet with especially strong action-predictiveness seems to have a slight advantage over the others. However, in absolute terms, performance still degrades significantly for DragonNet, as can be seen in the last row with Prec$_{\text{Ass.}}^{\hat{\pi}}$ as the y-axis. At maximal treatment effect bias, all models fail to suggest correct treatments completely, as a precision of 0.5 corresponds to random guessing. This further demonstrates that in settings with high relevant bias, counterfactual prediction is very difficult and inherently limited. This makes the need for models that are able to deal with as much bias as possible evident and shows that simple models like the S-Learner already fail for small amounts of bias. Another surprising finding is that for high $Y^1$-bias and treatment effect bias, the simple ActionNet outperforms all other models. Hence, in that case, it is better to simply learn the existing observational policy instead of trying to find a new one by estimating the treatment effect. In fact, DragonNet only seems to have an edge as soon as the ActionNet surpasses the other models. This implies that action predictive models are only better than balancing ones as soon as the models are not able to learn better decision making than the observational policy itself provides!

\paragraph{Biomarker Identification.} 
For AY-TCGA, we see 
in the last two rows in Fig.\ \ref{fig:tcga}
that if there is no relevant bias, i.e.\ the last two columns, 
all models, except for the ActionNet, are able to almost perfectly identify which features are true biomarkers. Since there is no mutual information between the treatment assignment and the outcomes, we also do not expect ActionNet to be able to identify any biomarkers. Under $Y^0$-bias, the predictive feature attribution performance correlates with the PEHE. Intuitively, this makes sense, because if a model is able to predict the correct outcomes and effects, it must have been able to pick up on important features.
For the prognostic feature score, however, the DragonNet outperforms the other Torch models. Here, DragonNet may be able to exploit the mutual information between treatment assignment and control outcome as an inductive bias for finding prognostic features. This principle is also evident in the setting with high effect bias (third column). DragonNet performs equally well to other Torch models in terms of PEHE, but outperforms them in predictive attribution for high bias scales. $Y^1$-bias has the most dramatic effect on predictive attribution. Here, TARNet and DragonNet identify only a small subset of the true predictive biomarkers and ActionNet surpasses all other Torch models for high bias. This stark difference 
between both types of outcome bias is probably 
because
the true predictive features are comprised only by $X^1$, since treatment (0) is chosen to represent a control. This is also reflected in that the counterfactual prediction for $Y^1$ is much worse for high $Y^1$-bias than for high $Y^0$-bias. 

\section{Conclusion}
This study investigates the complex relationship between biases in high-dimensional observational data and their impact on counterfactual prediction, treatment decision-making, and biomarker identification in precision medicine. A central contribution is the formalization of various biases 
based on mutual information, illustrating how these biases can influence model performance, particularly in high-dimensional datasets where traditional assumptions like overlap are often violated. A main finding is that not all biases negatively affect model performance and biomarker identification. Biases related to factors with minimal association with outcome mechanisms tend to have little impact on the accuracy of counterfactual predictions,
which is a positive message in the context of high-dimensional medical data.
Moreover, in settings of high treatment effect bias, corresponding to highly regulated clinical settings and "correct" observational treatment policy, there is a substantial gap between 
the useful information present in the treatment assignment policy, and the performance of all state-of-the-art counterfactual ML models, which all fail to exploit
this inductive bias. 
This shows 
the need for more research on developing counterfactual methods in these settings that can exploit this inductive bias instead of removing it.
Overall, these findings suggest that identifying and understanding the specific biases present in a dataset leads to more informed ML model selection and more reliable treatment decisions.
Additionally, using empirical potential outcomes from in-vitro cell line experiments offers a novel benchmark for evaluating counterfactual models, addressing the limitations of relying solely on semi-synthetic datasets. The results 
demonstrate the need for caution when interpreting performance on fully simulated data. 
The performed experiments and findings have
substantial 
implications for learning personalized treatment decisions from clinical observational data with AI and ML, 
where accurately predicting treatment outcomes in the presence of bias is crucial for developing effective personalized treatment plans. By appropriately developing modeling and adapting to biases in clinical data, more reliable and individualized treatment policies can be learned from high-dimensional multi-modal biological data, potentially leading to better patient outcomes in complex diseases like cancer. Future research should focus on developing AI and ML methods to assess and exploit present biases in real-world datasets, 
collecting and benchmarking datasets with real biological outcomes for evaluation, and advancing the theoretical understanding of performance bounds using these new insights.
%
In conclusion, this study helps to understand the role of treatment assignment bias in counterfactual prediction
and biomarker identification, paving the way for the development of more robust and accurate predictive algorithms crucial for personalized healthcare.


\cleardoublepage


\bibliography{references}

\appendix

\section{Additional Results}\label{apd:tcga}

\begin{figure*}[h]
    \centering
    \includegraphics[width=1\textwidth]{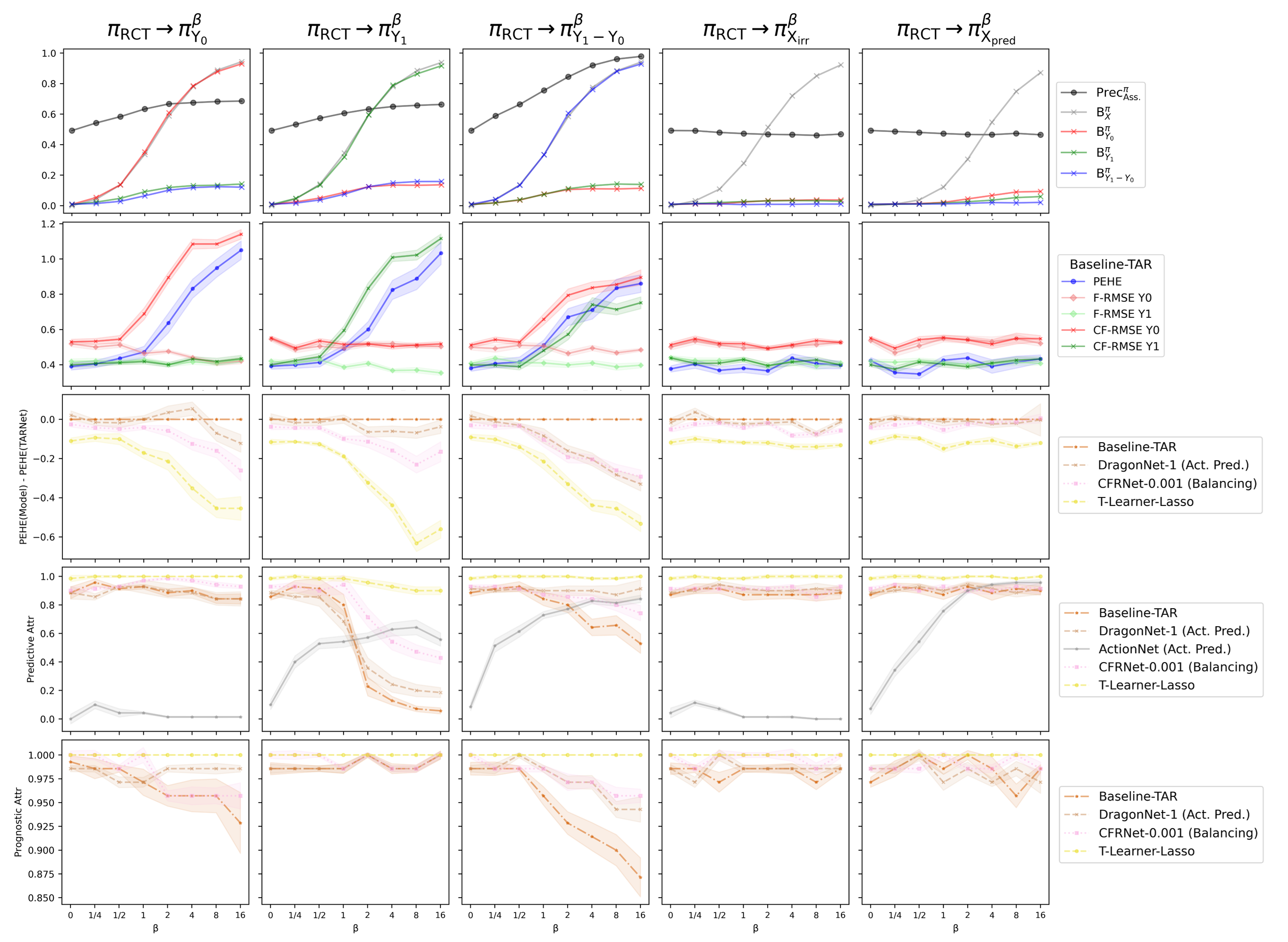} 
    \caption{Results for AY-TCGA. The experiment and the main findings are discussed in the main text.}
    \label{fig:tcga}
\end{figure*}

\begin{figure*}[h]
    \centering
    \includegraphics[width=1\textwidth]{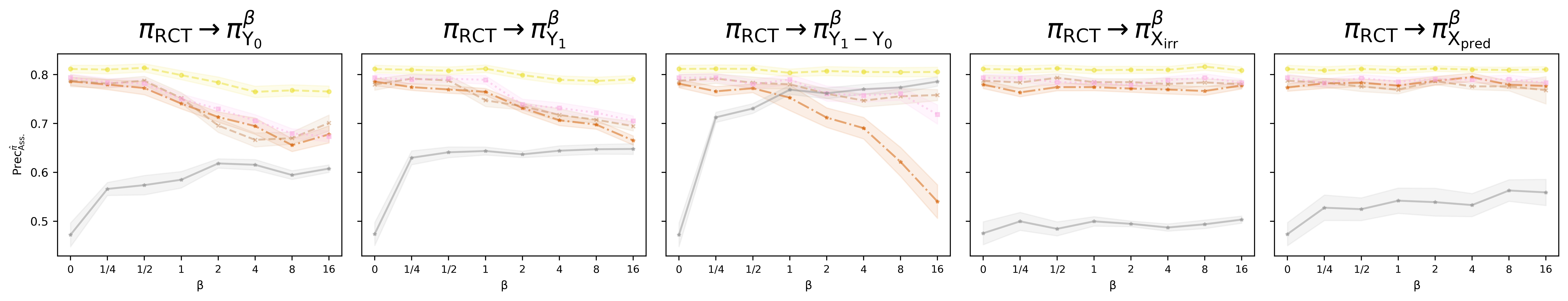} 
    \caption{Precision for AY-TCGA.}
    \label{fig:tcga2}
\end{figure*}

\begin{figure*}[ht]
    \centering
    \includegraphics[width=\textwidth]{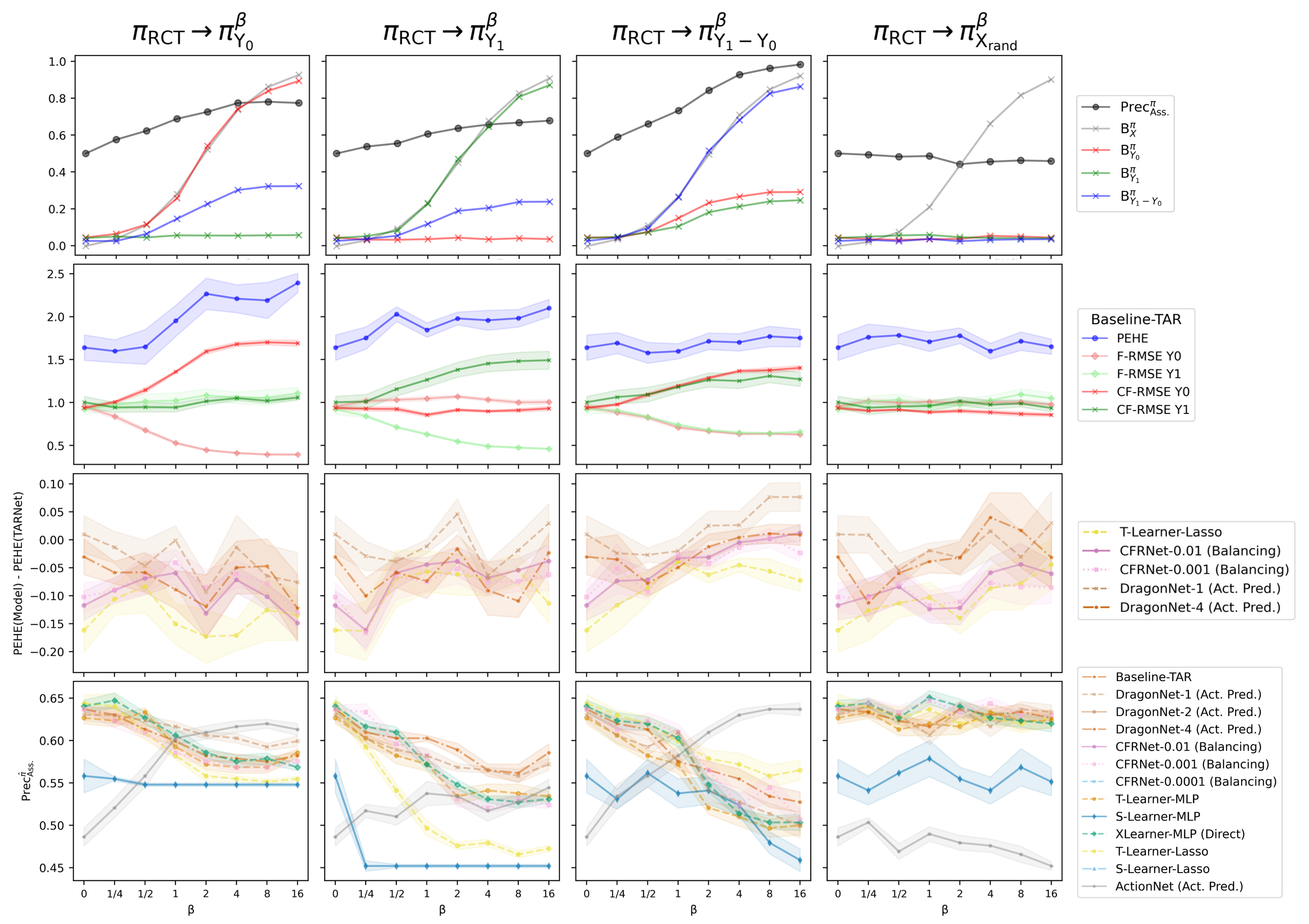} 
    \caption{Results for A-Drug.}
    \label{fig:drug}
\end{figure*}

This section presents additional experimental results.
\figureref{fig:tcga2} shows the precision results for A-TCGA, which were omitted from the results in the main text but discussed in Section \ref{sec:results}.
For the A-DRUG dataset, which includes drug response data derived from the DepMap project. The results for A-DRUG in \figureref{fig:drug} are consistent with the findings from the other datasets, such as A-CRISPR and AY-TCGA. Similar to those datasets, we observe that the degree of bias introduced by the observational policies significantly impacts the model's performance in predicting treatment effects and potential outcomes. Specifically, as the bias increases, factual prediction performance tends to improve slightly, while counterfactual prediction performance generally deteriorates, particularly in scenarios with high treatment effect bias. Overall, the results from the A-DRUG dataset reinforce the key findings of our study, confirming the importance of understanding and addressing bias in treatment effect estimation across different types of biological data. 

\FloatBarrier

\FloatBarrier

\begin{figure*}[t]
    \centering
\includegraphics[width=\textwidth]{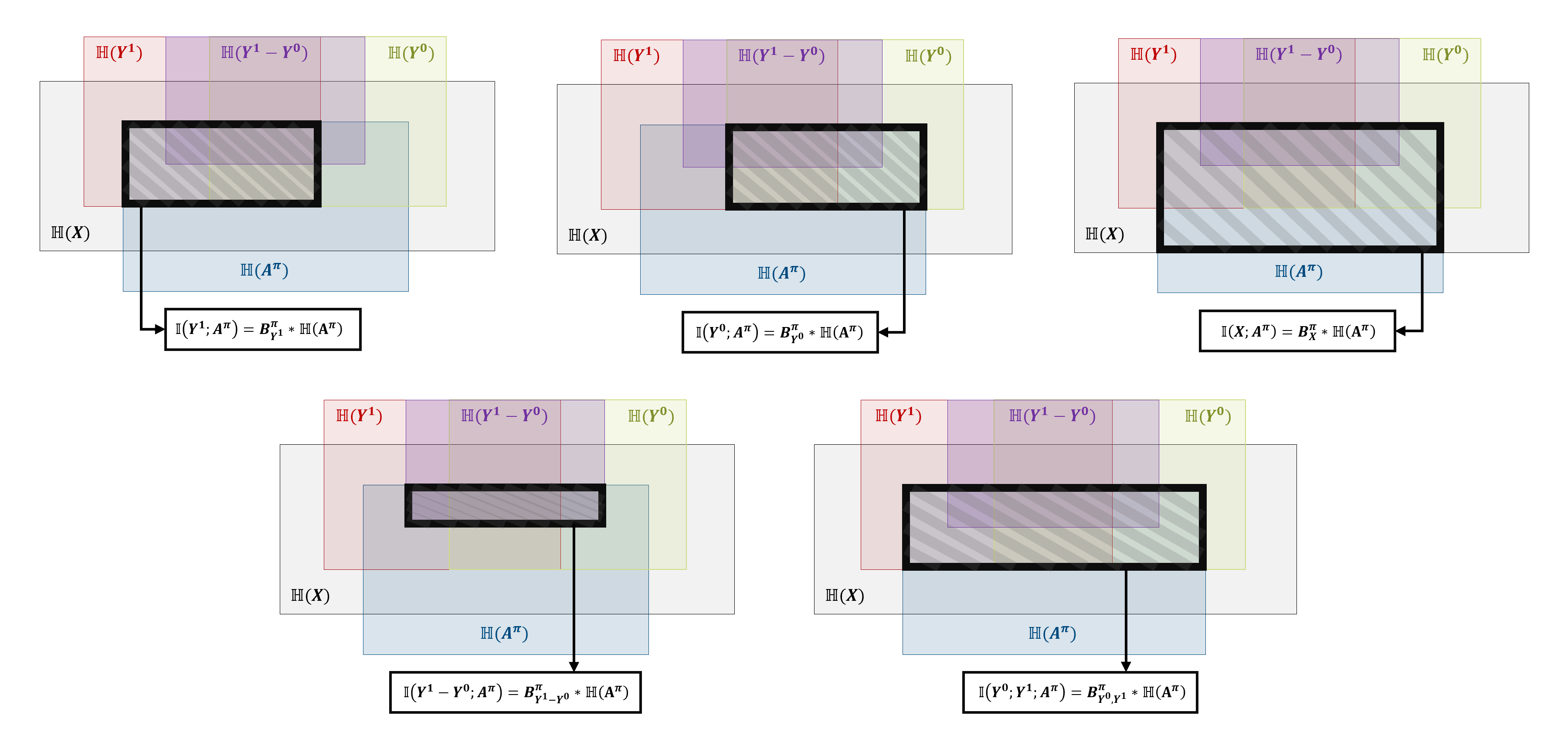} 
    \caption{Types of Bias.}
    \label{fig:biases}
\end{figure*}

\section{Proofs of Propositions}\label{apd:proofs} 
\subsection{Proof of Proposition \ref{prop:ineq}}
First, we prove that $B^\pi_X \geq B^\pi_{Y^0, Y^1}$. The approach is directly adopted from the proof for Proposition 1 by \citet{definingExpertise}. Under the non-confounding assumption, we have for all $Y^0,Y^1 \perp\!\!\!\perp A^{\pi} \mid X$, i.e. given $X$, we know that the outcomes are independent of treatment assignment. We proceed as follows:
\begin{align}
    B_{Y^0,Y^1}^\pi &= \frac{\mathbb{I}(A^\pi;Y^0,Y^1)}{\mathbb{H}[A^{\pi}]} \label{eq:1} \\
    &= 1 - \frac{\mathbb{H}[A^{\pi}|Y^0,Y^1]}{\mathbb{H}[A^{\pi}]} \label{eq:2} \\
    &\leq 1 - \frac{\mathbb{H}[A^{\pi}|Y^0,Y^1,X]}{\mathbb{H}[A^{\pi}]} \label{eq:3} \\
    &= 1 - \frac{\mathbb{H}[A^{\pi}|X]}{\mathbb{H}[A^{\pi}]} \label{eq:4} \\
    &= \frac{\mathbb{I}(A^\pi;X)}{\mathbb{H}[A^{\pi}]} \label{eq:5} \\
    &= B_{X}^\pi \label{eq:6},
\end{align}
where \eqref{eq:1} and \eqref{eq:6} follow from the definition of $Z$-bias, \eqref{eq:2} and \eqref{eq:5} by the definition of mutual information, \eqref{eq:3} holds since conditioning never increases entropy and \eqref{eq:4} follows from the conditional independence via the non-confounding assumption. Let $Z' \in \{Y^0,Y^1,Y^1-Y^0\}$, then second part of the inequality can be derived:
\begin{align}
    B_{Y^0,Y^1}^\pi &= \frac{\mathbb{I}(A^\pi;Y^0,Y^1)}{\mathbb{H}[A^{\pi}]} \nonumber \\ \nonumber
    &= 1 - \frac{\mathbb{H}[A^{\pi}|Y^0,Y^1]}{\mathbb{H}[A^{\pi}]}  \\
    &= 1 - \frac{\mathbb{H}[A^{\pi}|Y^0,Y^1,Z']}{\mathbb{H}[A^{\pi}]} \label{eq:3b} \\
    &\geq 1 - \frac{\mathbb{H}[A^{\pi}|Z']}{\mathbb{H}[A^{\pi}]} \label{eq:4b} \\
    &= \frac{\mathbb{I}(A^\pi;Z')}{\mathbb{H}[A^{\pi}]} \nonumber \\
    &= B_{Z'}^\pi \nonumber,
\end{align}
where \eqref{eq:3b} follows from the fact that $Z'$ is a deterministic function of $Y^0$ and $Y^1$ and therefore does not affect the entropy when conditioned on in addition to $Y^0$ and $Y^1$. \eqref{eq:4b} holds, since conditioning never increases entropy. Combining the two inequalities concludes the proof. See \figureref{fig:biases} for an intuitive understanding of these inequalities. The diagrams only depict one out of several possible relationships between the entropies. If the mechanisms do not introduce any noise, then the inequalities in \eqref{eq:bias_ineq} also hold for the respective entropies and further restricts the ordering of rectangles in the Venn diagram. 
\hfill \(\square\)
\subsection{Proof of Proposition \ref{prop:oa}} 
When $B_X^\pi = 1$, $ A^\pi $ is perfectly predictable from $ X $. This situation necessarily violates the overlap assumption, which requires that the probability $ P(A^\pi = 1 \mid X) $ to be strictly between 0 and 1 for all $ X $. Therefore, if $ B_X^\pi = 1$, the overlap assumption cannot be satisfied because there will be some values of $ X $ for which $ P(A^\pi = 1 \mid X) $ is either 0 or 1. See proof for Proposition 4 in \citet{definingExpertise} for a more rigorous approach. 
\hfill \(\square\)
\\
\\
Conversely, if the overlap assumption is violated, it does not necessarily imply that $ B_X^\pi = 1 $. Specifically, overlap is violated if $ P(A^\pi = 1 \mid X) $ equals 0 or 1 for some values of $ X $. This does not ensure that the mutual information is at its maximum because there might still be some uncertainty in $ A^\pi $ given $ X $ in other regions, preventing $ I(A^\pi; X) $ from reaching its theoretical maximum.
\\
\\
However, if the overlap assumption is violated across the entire support of $ X $, meaning that $ P(A^\pi = 1 \mid X) $ is 0 or 1 for every $ X $ within its support, then $ I(A^\pi; X) $ will indeed be maximized and hence $B_X^\pi = 1$.


\section{Bias vs. Expertise}\label{apd:bias_discussion}
\subsection{Terminology}
In the context of our work, we choose to use the term \textit{z-bias} rather than \textit{expertise} to describe the information-theoretic quantity proposed. For instance, \citep{definingExpertise} 
refer to it as expertise because, intuitively, if a doctor always assigns the correct treatment, this consistency signifies a high level of expertise. In such cases, the mutual information between the treatment assignment and the outcome would be large, reflecting a strong inductive bias that can be exploited for beneficial effects. However, the term \textit{expertise} falls short in scenarios where mutual information remains high, even when the doctor consistently assigns the wrong treatment. Although mutual information would still be maximized, labeling this as \textit{expertise} would be misleading since we would not consider a doctor an expert if their decisions are always incorrect. In that way, Prec$_{Ass.}^\pi$ may better describe the intended meaning of expertise. This is why we prefer the term \textit{bias}, which carries a more neutral or even negative connotation, appropriately highlighting the potential challenges this quantity introduces, such as making counterfactual predictions more difficult. Furthermore, the term \textit{ bias} allows us to reason about different types of this quantity, offering a more flexible framework to understand its implications in various clinical settings. 

\subsection{Results}
The \textit{Worst$\rightarrow$Expert} experiment performed by \citet{definingExpertise} for Figure 3 corresponds directly to increasing $\beta$ for $\pi_{X_{irr}^\beta}$ in \figureref{fig:tcga}. While their findings state that increasing irrelevant bias comes with a strong deterioration in performance, our results imply that the PEHE remains near constant, despite the increasing irrelevant bias. It has been confirmed with the authors that this discrepancy can be attributed to a bug in the implementation. The corrected experiment demonstrates a behaviour that is in agreement with the results in \figureref{fig:tcga}. 

\section{Data}\label{apd:data}

\subsection{Simulated Outcomes}
The potential outcomes are simulated as simple weighted linear combinations of randomly selected features with added Gaussian noise. To ensure that we can evaluate the performance of the prognostic and predictive biomarker identification, we simulate a true control group by setting $Z^0=0$ in \figureref{fig:fig2}. This dataset will be referred to as AY-TCGA, indicating that $A^\pi$, as well as $Y^0$ and $Y^1$ are fully simulated. Data were obtained from the TCGA Research Network: https://www.cancer.gov/tcga.

\subsection{Biological Outcomes}
\paragraph{PRISM repurposing drug screens.} \citep{DepMap2024_drug_screen}, performed on 877 cell lines according to \citet{Corsello2020} record the effect of hundreds of drugs on various cancer cell lines. The goal is to identify new therapeutic uses for existing drugs through a multiplexed approach. For our use case we focused on the measurements for two drugs, imatinib and az-628. Imatinib is a well-known tyrosine kinase inhibitor used mainly for chronic myeloid leukemia (CML) and gastrointestinal stromal tumors (GISTs). Az-628 is a selective inhibitor of RAF kinase, a key player in the MAPK/ERK signaling pathway, which is often involved in cancer development. As covariates we use RNA transcriptomics data, retaining the 200 most correlated features as covariates. 

\paragraph{CRISPR knock-out screens.} \citep{DepMap2021_crispr} were conducted on 1067 cell lines following the methodology described by \citet{behan2019prioritization}, and targeted a myriad of genes of interest. They provide critical insights into gene dependencies by knocking out specific genes to observe their effects on cancer cell viability. As the two treatment options in our setting, we used the knock-out of genes EGFR (Epidermal Growth Factor Receptor) and KRAS (Kirsten Rat Sarcoma Viral Oncogene Homolog), which are crucial in cell signaling pathways related to cancer, with mutations or overexpression frequently observed in various cancers. Again, the most correlated 200 RNA transcriptomics features were used as covariates for our analysis. We refer to these datasets as A-CRISPR and A-DRUG.

\section{Metrics}\label{apd:metrics}
\paragraph{PEHE.} The Precision in Estimation of Heterogeneous Effect (PEHE) is conventionally used for evaluating Conditional Average Treatment Effect (CATE) predictions. It is calculated as
\[
\textit{PEHE} = \sqrt{\frac{1}{n} \sum_{i=1}^n (\hat{\tau}(x_i) - \tau(x_i))^2},
\]
where \(\hat{\tau}(x_i)\) is the predicted treatment effect and \(\tau(x_i)\) is the true treatment effect for individual \(i\).

\paragraph{Precision.} The assignment precision of a policy $\textit{Prec}_{Ass.}^{\pi}$, inspired by \citet{curth2021really}, is another metric used in our study that measures the ratio of treatment options proposed correctly by the policy $\pi$. In this case, correct means that for a given patient, the proposed treatment option leads to a higher outcome than the alternative. For a trained model, $\hat{\pi}$, is the updated treatment assignment policy, which assigns according to $A^{\hat{\pi}} := d(X) = \mathbf{1}_{ \{\hat{\tau}(x) > 0\} }
$. We also considered the Area Under the Receiver Operating Characteristic Curve (AUROC) and the Area Under the Precision-Recall Curve (AUPRC). However, these metrics can be difficult to interpret due to the imbalance of positives and negatives, which is significantly impacted by the experimental knobs.

\paragraph{RMSE.} For the prediction of potential outcomes, we use the root mean square error (RMSE). This metric evaluates the accuracy of the predictions for both the factual and counterfactual outcomes of the test population. We differentiate between two different types of RMSE for both outcomes. The factual RMSE measures the error in outcome prediction for the treatment option to which a patient would have been assigned using the observational treatment policy. The counterfactual RMSE (CF RMSE) evaluates how well the outcomes were predicted for the alternative treatment option. 

\paragraph{Attribution Score.} Additionally, we evaluate the identification of biomarkers using the metric introduced by \citet{crabbé2022benchmarkingheterogeneoustreatmenteffect}. This metric assesses how well biomarkers predict treatment response heterogeneity:
\begin{equation}
    \text{Attr}_{pred} = \frac{1}{|\mathcal{D}_{\text{test}}|} \sum_{X \in \mathcal{D}_{\text{test}}} \frac{\sum_{i \in \mathcal{I}_{\text{pred}}} |a_i(\hat{\tau}, X)|}{\sum_{i=1}^{d} |a_i(\hat{\tau}, X)|},
    \label{eq:attr_pred}
\end{equation}
where $\mathcal{I}_{pred}$ is the set of predictive features and $a_i$ represents a specific attribution for the $i$-th feature of a given model in the context of an explainability method. Since, we only evaluate on AY-TCGA, which is simulated with a true control, i.e. $Z^0=0$ in \figureref{fig:fig2}-I, $\mathcal{I}_{pred}$ represents all features in $X^1$. Analogously, we also evaluate prognostic biomarker identification using Attr$_{prog}$.
In this case, we are utilizing SHAP (SHapley Additive exPlanations) values to quantify the contribution of each feature to the prediction \citep{lundberg2017unified}. 

\paragraph{Empirical Approximation of Z-Bias.}
We can rewrite $Z$-bias as follows:
\begin{equation}
\begin{aligned}
    B_Z^\pi &= \frac{\mathbb{I}(A^\pi;Z)}{\mathbb{H}[A^{\pi}]} \\
    &= \frac{\mathbb{H}[Z]-\mathbb{H}[Z|A^\pi]}{\mathbb{H}[A^\pi]} \\
    &\approx \frac{\Tilde{\mathbb{H}}[Z]-\Tilde{\mathbb{H}}[Z|A^\pi]}{\mathbb{H}[A^\pi]}
\end{aligned}
\end{equation}
To approximate the entropy terms including the continuous $Z$, we followed the discretization scheme from \citet{definingExpertise}. Using numpy, the support $\mathcal{Z}$ of $Z$ for a given dataset is automatically binned into $k$ bins: $\mathcal{Z} = \mathcal{Z}_1 \cup \dots \cup \mathcal{Z}_k$. Then the observations $z^i$ are discretized according to those bins. We can approximate the entropy terms with:
\begin{equation}
\begin{aligned}
    \Tilde{\mathbb{H}}[A^\pi] &= 
    \sum_{a \in \{0,1\}} \frac{\left|i : a_{i} = a\right|}{n} \log_{2} \frac{\left|i : a_{i} = a\right|}{n}
    \label{eq:placeholder} \\
    \Tilde{\mathbb{H}}[Z] &= 
    \sum_{j \in [k]} \frac{\left|i : z_{i} \in \mathcal{Z}_j \right|}{n} \log_{2} \frac{\left|i : z_{i} \in \mathcal{Z}_j\right|}{n} \\
    \Tilde{\mathbb{H}}[Z|A^\pi] &= \sum_{a \in \{0,1\}, j \in [k]} \frac{\left| \{i : a^i = a, z^i \in \mathcal{Z}_j\} \right|}{n} 
    \\
    & \quad\quad\quad\quad\log_2 \frac{\left| \{i : a^i = a, z^i \in \mathcal{Z}_j\} \right|}{\left| \{i : a^i = a\} \right|} \nonumber
\end{aligned}
\end{equation}

\section{Learners}\label{apd:learners}
In our experiments, we employ a variety of models, each utilizing distinct approaches to deal with the difficulties of counterfactual prediction. Here we first describe the three axes along which we compare different models and then provide a short description of the selected models.  See Table \ref{tab:model_comparison} for an overview of the classification of the learners according the three axes discussed in the main text.

\subsection{High-Level Classification of Models}
\paragraph{Linear vs. Nonlinear.} Nonlinear models, particularly neural networks, are hypothesized to better manage complex biases arising from nonlinear data-generating mechanisms. They adapt more flexibly to complex patterns in data, potentially capturing intricate relationships within high-dimensional biological datasets. In contrast, linear models are more suited to address simpler biases and might offer better interpretability when relationships are straightforward, but may underperform in scenarios requiring modeling of more complex dependencies.

\paragraph{Action-Predictive vs. Balancing.} Action-predictive models focus on learning the treatment selection mechanism itself, making them aware of the existing bias in treatment assignment. By understanding the selection process, these models can leverage the inherent biases (such as physician expertise reflected in treatment decisions) to make more informed predictions. These models can potentially exploit inductive biases beneficially, although it might only be effective when the existing treatment effect bias is not so severe that it already degrades prediction quality. Balancing models, on the other hand, aim to adjust or "balance" the treatment groups to minimize the influence of covariate shift, effectively "removing" bias. This may offer robustness in situations where treatment selection bias is high, leading to stable prediction performance across groups. Some models can exhibit neither balancing nor action-predictive behavior or both at the same time. So in principle, one could consider introducing two axes instead.

\paragraph{Direct vs. Indirect.} This axis distinguishes models by whether they predict treatment effects directly or indirectly. Direct models estimate treatment effects outright, while indirect models first predict counterfactual outcomes and then calculate treatment effects by subtracting these predictions. Indirect methods might be more resilient in settings with strong treatment effect bias but minimal outcome bias, as seen in scenarios like Toy3 (referenced in Appendix). Here, indirect models benefit by focusing on outcome predictions, potentially mitigating the direct impact of treatment effect bias on their estimations.
\begin{table*}[h]
\centering
\begin{tabular}{|l|c|c|c|}
\hline
\textbf{Model}                  & \textbf{Linear/Nonlinear} & \textbf{Action-Predictive/Balancing} & \textbf{Direct/Indirect} \\ \hline
S-Learner (Lasso)               & Linear                    & -                            & Indirect                   \\ \hline
S-Learner (Torch)               & Nonlinear                 & -                            & Indirect                   \\ \hline
T-Learner (Lasso)               & Linear                    & -                            & Indirect                   \\ \hline
T-Learner (Torch)               & Nonlinear                 & -                            & Indirect                   \\ \hline
X-Learner (Lasso)               & Linear                    & -                            & Direct                   \\ \hline
X-Learner (Torch)               & Nonlinear                 & -                            & Direct                   \\ \hline
TARNet                          & Nonlinear                 & -                            & Indirect                 \\ \hline
CFRNet                          & Nonlinear                 & Balancing                            & Indirect                 \\ \hline
DragonNet                       & Nonlinear                 & Action-Predictive (\& Balancing)                   & Indirect                 \\ \hline
ActionNet                       & Nonlinear                 & Action-Predictive                    & -                   \\ \hline
\end{tabular}
\caption{Categorization of models.}
\label{tab:model_comparison}
\end{table*}

\subsection{Description of Selected Models}
\paragraph{S-Learner and T-Learner.}
The \textit{S-Learner} (Single-Learner) approach involves combining all the data into a single predictive model that includes the treatment indicator as one of the covariates. This model is then used to estimate potential outcomes for different treatment groups by evaluating the model's output with different values of the treatment indicator. The \textit{T-Learner} (Two-Learner), on the other hand, trains separate models for each treatment group. The outcomes for each group are estimated independently, and the treatment effect is determined by comparing these estimates. Both S-Learner and T-Learner \citep{kunzel2019metalearners} are foundational and commonly used baselines in treatment effect estimation, offering simplicity and interpretability, though they may struggle in cases of covariate imbalance or heterogeneous treatment effects.

\paragraph{X-Learner.}
The \textit{X-Learner} \citep{kunzel2019metalearners} acts as an example for a direct learner. This means that it directly predicts the CATE, without first estimating POs. The model employs a two-step procedure: first, it estimates the potential outcomes for each treatment group separately. In the second step, these initial estimates are refined by exploiting the fact that treatment effects may be more accurately learned in one group and then transferred to the other group. This transfer mechanism allows the X-Learner to reduce bias and variance, improving the accuracy of CATE estimation. 

\paragraph{TARNet.}
\textit{TARNet} (Treatment-Agnostic Representation Network) \citep{shalit2017estimating} is a model within the family of representation learning approaches aimed at improving CATE estimation. See \citet{curth2021nonparametricestimationheterogeneoustreatment} for a description of how it relates to other representation learning approaches. \textit{TARNet} constructs a shared representation for all units (regardless of their treatment status) through a deep neural network, which is agnostic to the treatment assignment. This shared representation is then fed into separate heads that predict outcomes for each treatment group. The central idea behind TARNet is to create a representation that captures essential information about the units while minimizing the influence of treatment assignment, thereby enabling the model to generalize well across different treatment groups.

\paragraph{CFRNet.}
\textit{CFRNet} (Counterfactual Regression Networks) \citep{shalit2017estimating} is a model that belongs to the class of balancing methods. These models are designed to minimize bias in treatment effect estimation by reducing covariate distribution differences between treated and control groups. CFRNet achieves this by learning a latent representation of the data in which treated and control units are more similar. This alignment in the representation space allows for more accurate comparisons between the groups, thereby facilitating improved treatment effect estimation. The underlying objective of CFRNet is to approximate counterfactual outcomes more effectively by balancing covariates, making it a powerful tool in observational studies where covariate imbalance is a significant concern. CFRNet-0.0001, CFRNet-0.001 and CFRNet-0.01 increasingly strongly penalize the distance between the representation spaces.

\paragraph{DragonNet.}
The \textit{DragonNet} \citep{shi2019adapting} is an action-predictive model that incorporates a unique feature—a propensity score prediction head—within its network architecture. This head is designed to predict the probability of treatment assignment (i.e., the propensity score) while simultaneously estimating potential outcomes. By explicitly modeling the treatment assignment mechanism, DragonNet leverages the inductive bias inherent in the observed treatment decisions. This dual objective aims to enhance the model's ability to make treatment recommendations that align closely with the observed actions, potentially improving policy evaluation and treatment assignment accuracy. DragonNet-1, DragonNet-2 and DragonNet-4 increasingly strongly penalize the propensity head loss.

\paragraph{ActionNet.}
We introduce the \textit{ActionNet} to represent an extreme and trivial variant of action-predictive models. Unlike models that focus on estimating potential outcomes, ActionNet is solely concerned with predicting treatment decisions based on the observed selection policy. It trains a propensity model directly on the observed treatment assignments. The goal is to effectively mimic the existing observational policy without explicitly modeling the potential outcomes. This approach maximally exploits the existing biases in treatment decisions. Inherently, the performance of this model can only be as good as the observational policy. 

\paragraph{Balancing vs. Action-Predictive.}
The primary distinction between balancing models such as CFRNet and action-predictive models like DragonNet and ActionNet lies in their approach to handling bias \citep{definingExpertise}. Balancing models focus on mitigating bias by aligning the covariate distributions between treatment groups. This alignment allows for more fair and accurate comparisons of outcomes across these groups, which is critical in settings where treatment assignment is non-random and covariate imbalance may confound the estimated treatment effects. In contrast, action-predictive models take advantage of the inductive biases present in the observed treatment decisions. By modeling the treatment assignment process directly, these models aim to improve decision-making by leveraging the information embedded in the historical treatment patterns. However, this approach can be risky, as it relies heavily on the correctness and appropriateness of the observed treatment policy, potentially leading to biased recommendations if the underlying policy is flawed.

\paragraph{Model-Specific Limitations for Evaluation.}
Since XLearner and ActionNet do not estimate any potential outcomes, PO prediction cannot be evaluated for them. The ActionNet only predicts a decision and can therefore also not be evaluated on treatment effect prediction.
\\
\\
In summary, balancing models aim to reduce bias by ensuring fair comparisons across treatment groups, while action-predictive models leverage the bias inherent in the data to improve decision-making, with each approach offering distinct advantages depending on the context of the treatment effect estimation problem.

\begin{figure*}[h]
    \centering
    \includegraphics[width=0.8\textwidth]{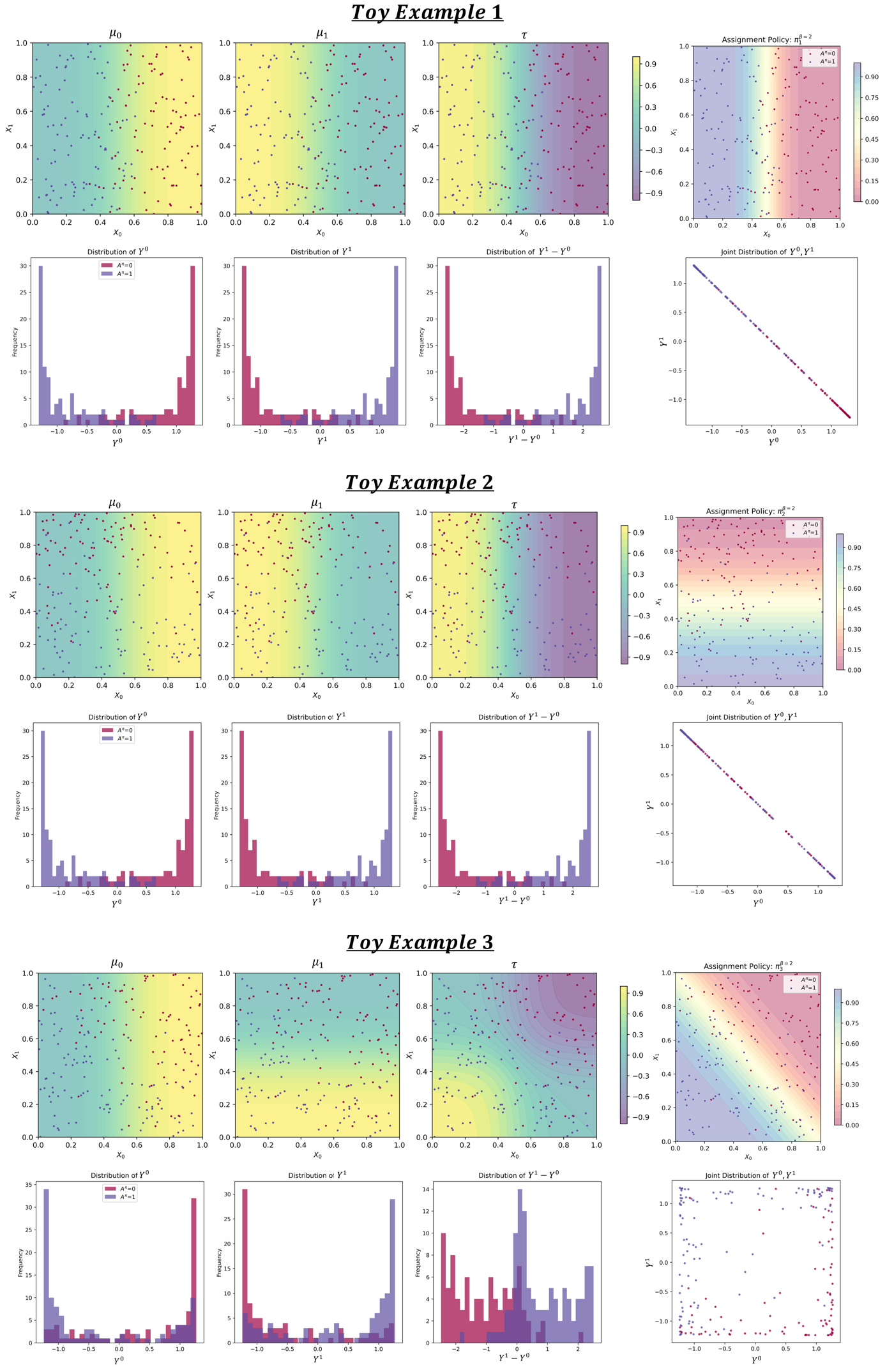} 
    \caption{Visualization of Toy Examples 1-3.}
    \label{fig:toy1_3}
\end{figure*}

\begin{figure*}[h]
    \centering
    \includegraphics[width=0.8\textwidth]{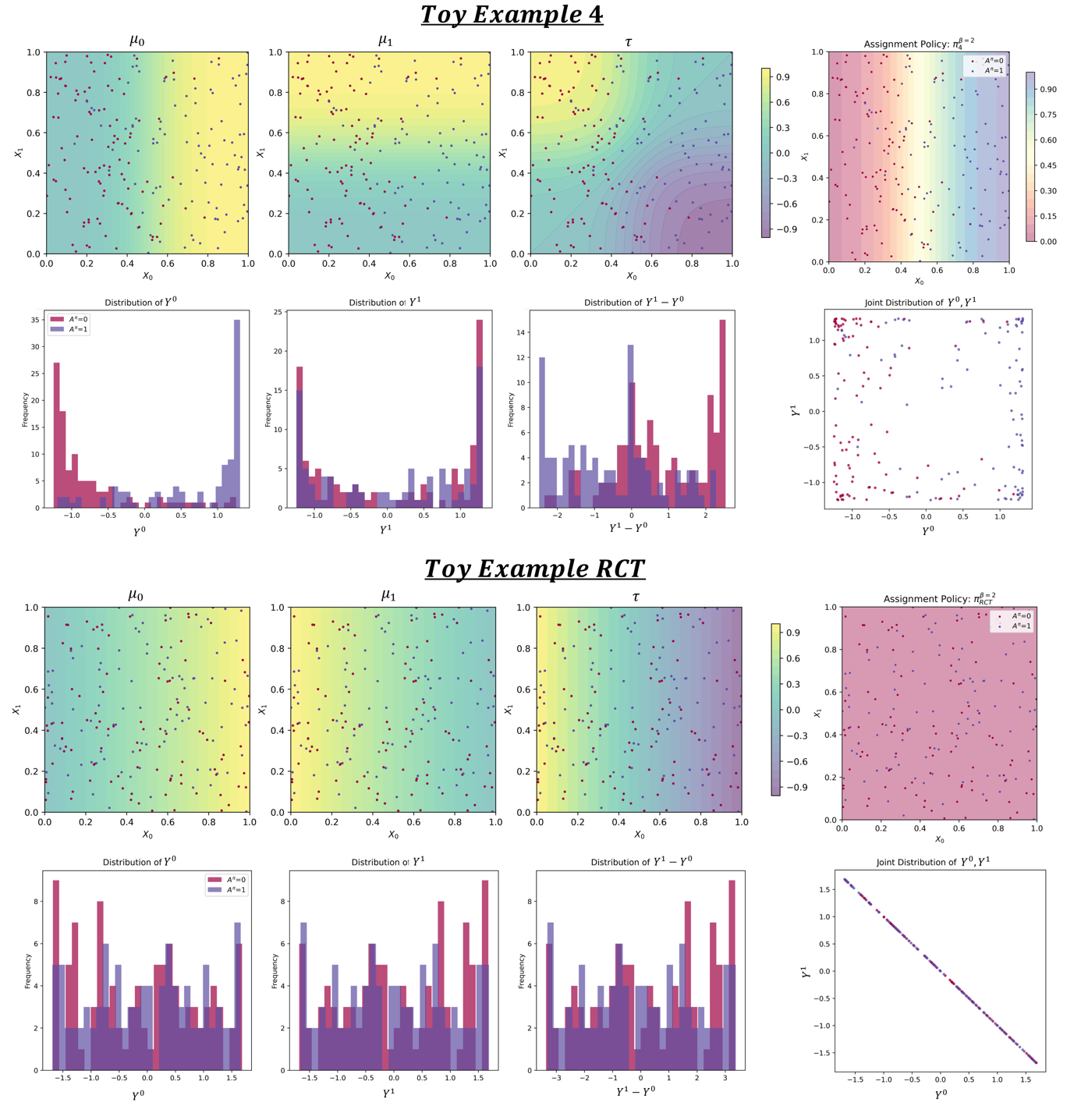} 
    \caption{Visualization of Toy Examples 4 and RCT.}
    \label{fig:toy4_rct}
\end{figure*}



\section{Toy Examples}\label{apd:toy_examples}
\subsection{Construction}
\figureref{fig:toy1_3,fig:toy4_rct} depict several toy examples, constructed to demonstrate the effect of selection bias. The underlying DGPs are special cases of the structural causal model in \figureref{fig:fig2}. All simulated outcomes follow the structure $Y^a = f_{nl}(f_i^a(x_0,x_1))$, where $f_{nl}(\cdot) = \frac{1}{1+\exp(-10*(\cdot-0.5))}$ is a logistic-like nonlinearity. The policies are designed to be proportional to different quantities and are computed similarly to \equationref{eq:policy}.

\paragraph{Toy1.}
\begin{equation}
\begin{aligned}
    f_1^0(x_0,x_1) &:= x_0 \\
    f_1^1(x_0,x_1) &:= 1-x_0 \\
    \pi^\beta_1(x_0,x_1) &\sim 1-x_0 \\
\end{aligned}
\end{equation}
This toy example should show why the covariate shift, visible in the outcome and treatment effect plots, leads to worse performance in estimating the treatment effect and outcomes as compared to the RCT setting. For PO prediction, we hypothesize that factual prediction performance will increase slighly, since the covariate shift makes the support of the factual part of the outcome surface smaller and thereby reduces the variance of its value range, making it easier to learn. Counterfactual prediction, however, will deteriorate, since the counterfactual part of the outcome surface is underrepresented in the training cohort and the model will fail to extrapolate correctly to those regions. The lack of overlap of the outcome and treatment effect value ranges is visualized by the histograms and directly relates to the respective types of bias. This toy example leads to the sharp separation of the two treatment hues of the histogram bars, indicating high bias.

\paragraph{Toy2.}
\begin{equation}
\begin{aligned}
    f_2^0(x_0,x_1) &:= x_0 \\
    f_2^1(x_0,x_1) &:= 1-x_0 \\
    \pi^\beta_2(x_0,x_1) &\sim 1-x_0 \\
\end{aligned}
\end{equation}
Here, the only thing different from Toy1, is that the treatment policy depends on a different dimension than the outcomes. As for Toy1, with high $\beta$, the policy, will violate the overlap assumption as the observable bias goes approaches 1. Due to the independence of the assignment and outcome mechanisms, however, all other biases will remain 0. The histograms nicely illustrate how there is full overlap of the patient groups with respect to the outcome value ranges, visualizing the lack of bias.

\paragraph{Toy3.}
\begin{equation}
\begin{aligned}
    f_3^0(x_0,x_1) &:= x_0 \\
    f_3^1(x_0,x_1) &:= 1-x_1 \\
    \pi^\beta_3(x_0,x_1) &\sim 1-x_0-x_1 \\
\end{aligned}
\end{equation}
This example is constructed to lead to low treatment outcome biases, but high treatment effect bias. It should demonstrates how such constructed examples can be used to test hypotheses about properties of different kinds of models. Here, for instance, we speculate that indirect learners, estimating the treatment effect via outcome predictions, will show better performance than direct learners, which estimate the treatment effect without having to estimate potential outcomes. 

\paragraph{Toy4.}
\begin{equation}
\begin{aligned}
    f_4^0(x_0,x_1) &:= x_0 \\
    f_4^1(x_0,x_1) &:= x_1 \\
    \pi^\beta_4(x_0,x_1) &\sim x_0 \\
\end{aligned}
\end{equation}
This setting is constructed to exhibit high $Y^0$-bias, but no $Y^1$-bias. The idea is to demonstrate that different types of bias can have fundamentally different effects on prediction. Further, we want to evaluate whether the high treatment effect bias can also be exploited as inductive bias, as proposed by \citet{definingExpertise}. 

\begin{figure*}[h]
    \centering
    \includegraphics[width=\textwidth]{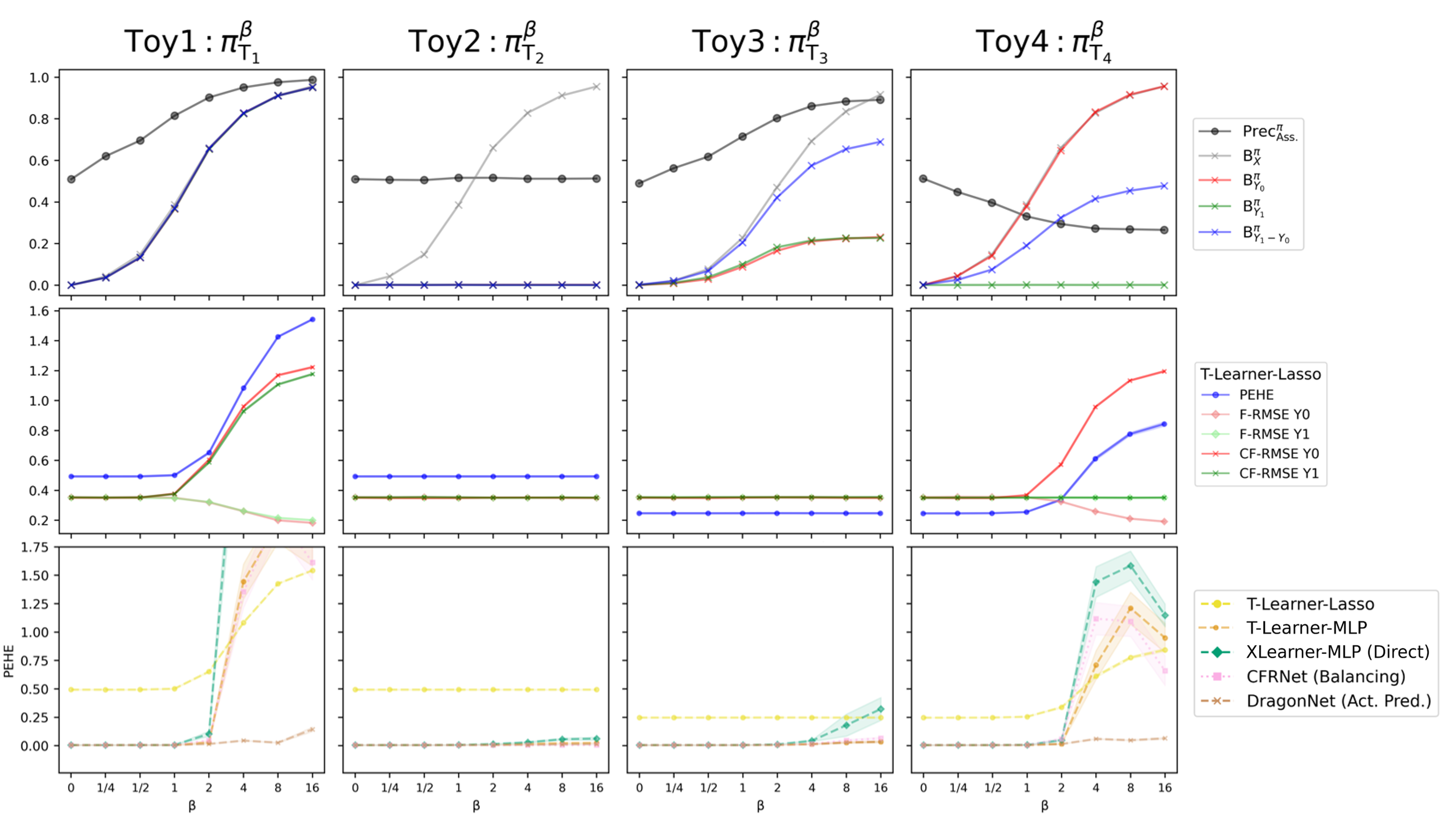} 
    \caption{Results for toy examples 1-4.}
    \label{fig:toy}
\end{figure*}
\subsection{Results on Toy Examples}
The first two rows of \figureref{fig:toy} generally show that the biases behave as constructed and that the change in performance in terms of PEHE and RMSE confirm expectations. 

\paragraph{Toy1.} For Toy1, where all types of bias increase with increasing bias scale $\beta$, counterfactual prediction for both outcomes and treatment effect prediction become significantly worse for the T-Learner-Lasso. Factual prediction, on the contrary, improves slightly. In the third row we observe that for low bias, neural-network based models clearly outperform the linear TLearner. This is not surprising, as the treatment effect surface is constructed to be nonlinear. However, with increasing bias, it seems that the T-Learner-Lasso and more significantly so, the Dragon-Net, suffer less from bias than all other models. The Dragon-Net may indeed by able to exploit the strong inductive bias captured by the obersvational policy. The amount of inductive bias is represented by Prec$_{Ass.}^\pi$. The more correct decisions the obervational policy already made, the more inductive bias may be available to exploit. Note, that also an obervational policy which always decides incorrectly may contain inductive bias. The results for 

\paragraph{Toy2.} Toy2 provides an example for when the violation of the overlap assumption has little to no effect on the performance for all models. This in support of the hypothesis that not all types of bias are bad for counterfactual prediction performance. 

\paragraph{Toy3.} Here, the results show that only the direct Xlearner experience a performance degradation due to the introduced treatment effect bias. This may be explained by the fact, that indirect learners only estimate potential outcomes and not the treatment effect. Therefore, a situation where there is little outcome bias but large treatment effect bias may be advantageous for indirect learners, compared to direct learners.

\paragraph{Toy4.} The plot in the second row for Toy4, nicely demonstrates that outcome prediction only deteriorates if there is bias with respect to that outcome. As for Toy1, it seems, Dragon-Net significantly outperforms the other models. While there is less inductive bias than for Toy1, it may still be able to exploit it to some extent.


\end{document}